\documentclass[10pt,twocolumn,letterpaper]{article}

\usepackage[pagenumbers]{iccv} 

\newcommand{\cmark}{\ding{51}}%
\newcommand{\xmark}{\ding{55}}%

\usepackage{tabularx}
\usepackage{multirow}
\usepackage{arydshln}
\usepackage{pifont}
\usepackage{makecell}
\usepackage[normalem]{ulem}
\usepackage[margin=1in]{geometry}
\usepackage{booktabs}    
\usepackage{xcolor}      
\usepackage{multirow}
\usepackage{glossaries}

\newacronym{our}{GECKO}{\textbf{G}igapix\textbf{E}l Vision-\textbf{C}oncept \textbf{K}nowledge C\textbf{O}ntrastive Pretraining}

\newacronym{si}{SI}{Self-Interpretable}
\newacronym{mil}{MIL}{Multiple Instance Learning}
\newacronym{wsi}{WSI}{whole-slide image}
\newacronym{vlm}{VLM}{Vision Language Model}
\newacronym{llm}{LLM}{Large Language Model}
\newacronym{ssl}{SSL}{self-supervised learning}
\newacronym{fm}{FM}{foundational model}
\newacronym{vcm}{VCM}{Vision Concept Model}

\definecolor{iccvblue}{rgb}{0.21,0.49,0.74}
\usepackage[pagebackref,breaklinks,colorlinks,allcolors=iccvblue]{hyperref}
\usepackage{subcaption}  
\usepackage{graphicx}

\title{\includegraphics[height=0.7cm]{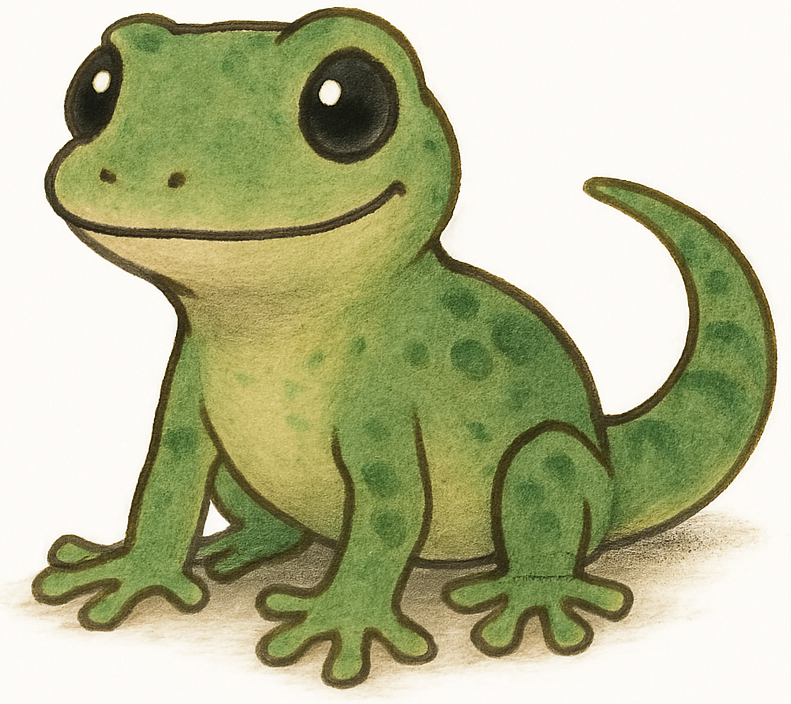}\hspace{0.2cm}GECKO: Gigapixel Vision-Concept Contrastive Pretraining in Histopathology}

\usepackage[para]{footmisc}

\newcommand*\samethanks[1][\value{footnote}]{\footnotemark[#1]}

\author{Saarthak Kapse$^{1}$\thanks{Co-first authors,} , Pushpak Pati$^{3}$\samethanks[1] , 
Srikar Yellapragada$^1$\thanks{Co-second authors} , Srijan Das$^2$\samethanks[2] , 
Rajarsi R. Gupta$^1$ \\ Joel Saltz$^1$, Dimitris Samaras$^1$, Prateek Prasanna$^1$ \\ \\
$^1$Stony Brook University, USA \hspace{0.2cm} $^2$UNC Charlotte, USA \hspace{0.2cm} $^3$Independent Researcher
}

\begin{document}
\maketitle
\begin{abstract}
\vspace{-0.5cm}

Pretraining a Multiple Instance Learning (MIL) aggregator enables the derivation of Whole Slide Image (WSI)-level embeddings from patch-level representations without supervision. While recent multimodal MIL pretraining approaches leveraging auxiliary modalities have demonstrated performance gains over unimodal WSI pretraining, the acquisition of these additional modalities necessitates extensive clinical profiling. This requirement increases costs and limits scalability in existing WSI datasets lacking such paired modalities. To address this, we propose Gigapixel Vision-Concept Knowledge Contrastive pretraining (GECKO), which aligns WSIs with a Concept Prior derived from the available WSIs. First, we derive an inherently interpretable concept prior by computing the similarity between each WSI patch and textual descriptions of predefined pathology concepts. GECKO then employs a dual-branch MIL network: one branch aggregates patch embeddings into a WSI-level deep embedding, while the other aggregates the concept prior into a corresponding WSI-level concept embedding. Both aggregated embeddings are aligned using a contrastive objective, thereby pretraining the entire dual-branch MIL model. Moreover, when auxiliary modalities such as transcriptomics data are available, GECKO seamlessly integrates them. Across five diverse tasks, GECKO consistently outperforms prior unimodal and multimodal pretraining approaches while also delivering clinically meaningful interpretability that bridges the gap between computational models and pathology expertise. Code is made available at \href{https://github.com/bmi-imaginelab/GECKO}{github.com/bmi-imaginelab/GECKO}

\end{abstract}
    
\vspace{-0.4cm}

\section{Introduction}
\label{sec:intro}

\begin{figure}[!ht]
    \centering
    \includegraphics[width=1\linewidth]{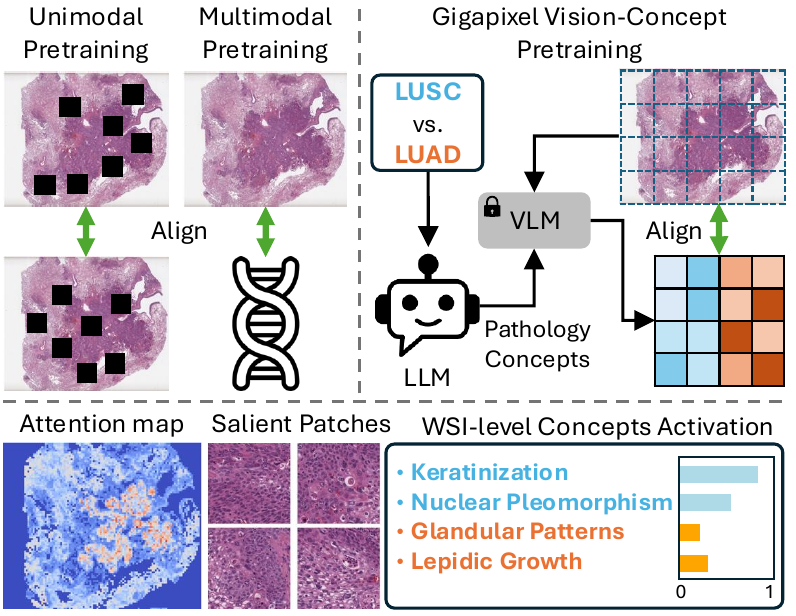}
    \caption{Unlike conventional unimodal or multimodal pretraining of WSI-level MIL aggregators, GECKO aligns a WSI with an interpretable Concept Prior derived from the WSI and task-relevant pathology concepts. Alongside downstream unsupervised and supervised performance benefits, GECKO provides WSI-level pathologist-friendly interpretable descriptors.}
    \label{fig:gecko_teaser}
    \vspace{-0.4cm}

\end{figure}

There has been a surge in \glspl{fm} in the histopathology domain~\cite{gigapath,uni,virchow}, scaling both model sizes and the number of \glspl{wsi}, which has led to significant improvements in downstream task performance. However, since \glspl{wsi} are gigapixel in nature, most of these models rely on decomposing a \gls{wsi} into small patches and encoding them individually into an embedding space. Consequently, unlike in natural imaging, to use existing patch-level \glspl{fm} for \gls{wsi}-level analysis, supervisory signals are needed to aggregate patch embeddings into slide-level embeddings; a process that is prevalently achieved by supervised training of a \gls{mil}-based aggregator~\cite{abmil,clam}.

To overcome this requirement of supervisory signals from human annotations, recent works~\cite{lazard2023giga,gigapath} have explored using a contrastive objective to pretrain the \gls{mil} aggregator on patch embeddings extracted from \glspl{wsi}. This approach enables the derivation of slide-level embeddings in an unsupervised manner. However, this pretraining paradigm faces two distinct challenges that are particularly critical to address in real-world applications. (1) the need for additional modalities to facilitate effective MIL pretraining at the \gls{wsi}-level, and (2) the lack of pathologist-friendly interpretability in the pretrained model's aggregated \gls{wsi}-level embeddings. Below, we describe these challenges in further detail.

\underline{Challenge 1: Need for Additional Modality.} Recent studies~\cite{jaume2024transcriptomics,MADELEINE} have raised concerns about unimodal pretraining of \gls{mil} solely on \gls{wsi} data, noting a tendency to overfit on staining artifacts instead of capturing biologically relevant features. This issue arises because the intra-slide invariance objective limits the diversity of the training signal, thereby amplifying the impact of staining variations between training and test distributions. In natural image analysis, integrating an auxiliary modality, such as textual information, into image pretraining serves as an effective strategy to enhance model robustness by mitigating overfitting to spurious visual correlations~\cite{clip}. Along these lines, recent computational pathology studies have explored multimodal contrastive pretraining, wherein the H\&E \gls{wsi} modality (using \gls{mil} aggregator) is aligned with paired transcriptomics~\cite{jaume2024transcriptomics} or with \gls{wsi} stained using IHC markers~\cite{MADELEINE}, demonstrating superior performance over unimodal pretraining. Nonetheless, their dependency on extra modalities for effective \gls{wsi}-level pretraining is \textit{limiting}, due to the relatively smaller size of paired modality datasets, the high cost of acquiring them, and the challenges associated with data harmonization and standardization across multiple sources.

\underline{Challenge 2: Limited Interpretability.} Though both unimodal and multimodal pretraining methods for MIL aggregators bypass the need for \gls{wsi}-level labels, during testing they yield only an aggregated \gls{wsi}-level embedding and corresponding patch attention scores (as \gls{mil} is often attention-based~\cite{abmil}) for a \gls{wsi}. Since these deep embeddings are inherently non-interpretable~\cite{simil,rudin2019stop}, the interpretability of such models is \textit{limited} to highlighting salient regions without revealing the key pathology concepts and insights that drive predictions. In contrast, when diagnosing a cancer patient's H\&E \gls{wsi}, a pathologist not only annotates the salient regions but also explains the diagnosis by identifying key visually discriminative pathology concepts, such as glandular or lepidic growth patterns in lung adenocarcinoma, or keratinization patterns in lung squamous cell carcinoma.

To this end, we ask: \textit{Can we pretrain an effective WSI-level MIL aggregator without an auxiliary data modality, one that also provides \gls{wsi}-level embeddings interpretable by pathologists?}

The answer is yes. In this paper, we propose the \textit{first effective \gls{wsi}-level pretraining solution}, that (1) does not require additional clinical profiling (e.g., RNA sequencing) for collecting paired modalities, and (2) provides expert-interpretable \gls{wsi}-level embeddings (see Figure~\ref{fig:gecko_teaser}).  
To this end, we first computationally derive a \gls{wsi}-level Concept Prior and then employ a dual-branch \gls{mil} model to align the \gls{wsi} with this Concept Prior.

\textbf{First}, we propose to derive a new pathology concept-driven prior from H\&E \gls{wsi} that is capable of providing task-specific discriminative signal for pretraining, while being inherently interpretable. 
This Concept Prior is a cosine similarity matrix computed in the vision-language embedding space.  
While the vision embeddings are obtained from WSI patches, the language embeddings are extracted from textual descriptions of pre-defined \textit{visually discriminative pathology concepts} pertinent to each class.  
Each value in the concept prior matrix encodes the activation of a concept in the corresponding patch, making it inherently interpretable.

\textbf{Second}, we propose \gls{our}, a pretraining method which employs a dual-branch \gls{mil} network comprising a \gls{wsi}-level deep encoding branch and a concept encoding branch.
On one hand, the deep encoding branch aggregates the deep features (patch features) into a \gls{wsi}-level deep embedding.  
On the other hand, the concept encoding branch aggregates the concept prior into a \gls{wsi}-level concept embedding through a linear mapping to preserve its interpretability~\cite{simil}.  
Finally, a contrastive objective~\cite{clip} aligns these modalities, 
thereby pretraining both \gls{mil} branches.  
While \gls{our} is robust and does not require additional modalities, it allows for seamless integration of additional modalities (e.g., transcriptomics data), if available, during pretraining.  
Such multimodal pretraining can always benefit from our concept prior.

The dual-branch \gls{mil} model pretrained with \gls{our} is evaluated in both unsupervised and supervised setup. We propose a pathologist-driven heuristic that leverages interpretable \gls{wsi}-level concept embeddings to classify a slide by identifying the class with the most dominantly activated concepts; this demonstrates dramatic performance improvement over previous unsupervised baselines~\cite{mizero}. In the supervised setting, we further enhance performance by combining \gls{wsi}-level deep embeddings with concept embeddings. Notably, \gls{our} surpasses existing unimodal pretraining methods when only the \gls{wsi} modality is available. Moreover, when additional modalities like gene data are included, \gls{our} integrated with gene data exceeds the performance of existing multimodal pretraining approaches~\cite{jaume2024transcriptomics,panther,titan}. In summary, our main contributions are:

\begin{itemize}
    \item We derive a task-specific, interpretable \textit{concept prior} from H\&E \gls{wsi}, which enables effective slide-level pretraining without requiring additional clinical profiling.
    
    \item Our dual-branch \gls{mil} pretrained with \gls{our} offers \gls{wsi}-level deep embedding and interpretable concept embedding, with the latter enabling unsupervised \gls{wsi} prediction via a pathologist-driven heuristic.

    \item \gls{our} allows seamless integration of auxiliary modalities, and consistently delivers state-of-the-art performance on five slide-level tasks while offering pathologist-friendly interpretability for its predictions.

\end{itemize}

\section{Related Work}
\label{sec:related}
\textbf{\gls{wsi}-level pretraining:} Recent advances in pretraining methods~\cite{gigapath,lazard2023giga,jaume2024transcriptomics,MADELEINE,titan,hipt} for \gls{mil} models in histopathology have greatly improved downstream tasks, especially in few-label settings. For example, Giga-SSL~\cite{lazard2023giga} uses contrastive learning on gigapixel slides with sparse convolutional layers, removing the need for manual annotations. TANGLE~\cite{jaume2024transcriptomics} incorporates gene expression profiles to enhance slide representation learning, achieving better few-shot performance than unimodal pretraining on \gls{wsi} data alone. TANGLE also introduced an unimodal pretraining approach (Intra) that relies solely on \gls{wsi} data; which is limited by an intra-slide invariance objective that may overfit to staining artifacts~\cite{jaume2024transcriptomics,MADELEINE}, resulting in lower performance than multimodal pretraining. MEDELEINE~\cite{MADELEINE} employs different slide stainings of the same tissue with a global-local cross-stain objective to learn comprehensive morphological features, showing the superiority of multimodal pretraining. TITAN~\cite{titan} employs large patch-level pretraining (8k $\times$ 8k) using DINO~\cite{dino,dinov2}, followed by contrastive pretraining with CoCa~\cite{coca} on a large corpus of paired pathology reports and synthetic captions.  PANTHER~\cite{panther} uses Gaussian mixture models to summarize patches into morphological prototypes. Though it is not strictly a pretraining approach, it facilitates unsupervised patch-to-slide level aggregation. Our method, pretrained with only \gls{wsi} data, significantly outperforms Intra and matches TANGLE. Moreover, when gene data is integrated, our method significantly surpasses previous multimodal approaches.

\noindent\textbf{Interpretability methods:} 
Previously, SI-MIL~\cite{simil} introduced the first self-interpretable~\cite{cbm, cem, barbiero2023interpretable, labo} method in histopathology that highlights salient regions in a \gls{wsi} and also quantifies each handcrafted feature's contribution to predictions. Recently, Concept MIL~\cite{conceptmil} addressed the limitations of handcrafted features by replacing them with vision-language based concept activation scores, thereby offering more pathologist-friendly interpretability. However, these self-interpretable methods in this domain are limited to supervised setting. In our \gls{mil} model, which we aim to pretrain, we employ a linear mapping aggregator motivated by SI-MIL to encode the derived concept prior, thereby preserving interpretability, and then align this linearly aggregated concept prior with the gigapixel \gls{wsi}. While other works~\cite{watawana2024hierarchical,javed2024cplip,zhou2024knowledge,jaume2021quantifying} have incorporated concepts to enhance pretraining in histopathology, they focus on patch-level pretraining. Thus, our method marks the first approach to leverage concepts for scaling to \gls{wsi}-level pretraining.

\section{Proposed Method}
\label{sec:method}

\begin{figure*}[!ht]
    \centering
    \includegraphics[width=1\linewidth]{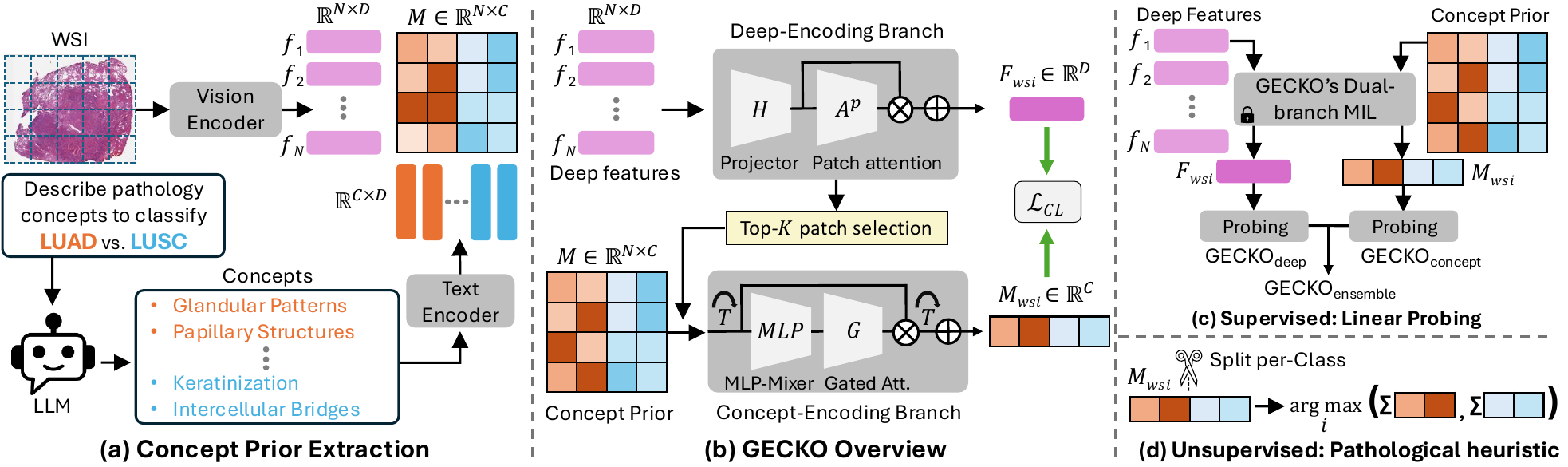}
    \caption{Overview of \gls{our}: (a) We start by extracting an interpretable, task-relevant Concept Prior from a \gls{wsi} using a \gls{llm} and a \gls{vlm}. (b) Next, we pretrain a dual-branch \gls{mil} by contrastively aligning \gls{wsi}-level deep and concept embeddings. (c) These embeddings can be used for supervised learning via linear probing. (d) Additionally, the concept embedding can be directly used for unsupervised learning using a pathologist-driven heuristic.}
    \label{fig:gecko_overview}
    \vspace{-0.3cm}

\end{figure*}

This section introduces \gls{our}, our gigapixel image-concept knowledge pre-training strategy, illustrated in Fig.~\ref{fig:gecko_overview}. \gls{our} pretrains a dual-branch \gls{mil} consisting of \gls{wsi}-level deep-encoding and concept-encoding branches, by contrastively aligning them. First, we detail the extraction of the concept prior from a \gls{wsi} in Sec~\ref{sec:concept_modality}. Then, we elaborate on the deep-encoding and concept-encoding branches in Sections Sec.~\ref{sec:deep_encoding_branch} and Sec.~\ref{sec:concept_encoding_branch}, respectively. Finally, we outline the \gls{our} pre-training and inference strategy in Sec.~\ref{sec:concept_guided_pretrain}.

\subsection{Concept Prior Extraction}
\label{sec:concept_modality}

The concept prior for a \gls{wsi} is extracted by using well-established task-specific pathology knowledge, ensuring that the resulting derived prior is both pathologist interpretable and provides appropriate discriminative signals for the task at hand. Note that, we focus on task-specific concepts instead of \gls{wsi}-specific labels.

First, we define the task-specific lexicon by leveraging a \gls{llm}, \eg, GPT-4. We define dedicated prompts to query the \gls{llm}, generating textual descriptions of \textit{visually discriminative pathology concepts} pertinent to each class $l \in L$ in the downstream task~\cite{conceppath}. These concept descriptions across $L$ classes are encoded into embeddings using the text encoder of a pre-trained \gls{vlm}~\cite{conch,huang2023visual,xiang2025vision,ikezogwo2023quilt,seyfioglu2024quilt}, denoted as $T \in \mathbb{R}^{C \times D}$, where $C$ is the total number of concepts and $D$ is the embedding dimension.
Next, each \gls{wsi} is divided into $N$ patches and each patch is encoded into a feature vector $f_i \in \mathbb{R}^{D}$ using the vision encoder from the same pre-trained \gls{vlm}, resulting in $F \in \mathbb{R}^{N \times D}$.
Given that both the text and image embeddings reside in a shared embedding space, we compute the pairwise cosine similarity between patch features $F$ and concept description embeddings $T$, yielding a concept matrix $M \in \mathbb{R}^{N\times C}$. $M$ serves as the concept prior for \gls{our}. Each element in $M$ quantifies the activation level of a specific concept within a patch, providing an interpretable mapping in the language of pathology. Furthermore, since we select a set of distinct concepts for each class, the concept prior matrix $M$ inherently encodes a task-specific discriminative signal. In particular, the salient regions of a \gls{wsi} belonging to a given class exhibit higher activation levels for the concepts associated with that class. Note that, although downstream classes may share some concepts, we only select those that are most visually unique to each class to provide a clear discriminative signal for pretraining.

To highlight, this concept prior extraction can be fully automated using LLMs, unlike other clinical modalities such as gene sequencing, pathology reports, and staining markers which require significant effort.

\subsection{WSI-level Deep-Encoding Branch}
\label{sec:deep_encoding_branch}

The \gls{wsi}-level deep-embedding is learned via a \gls{mil} model.
The patch-level deep features $f_i \in F$ from a \gls{wsi}, in Sec. \ref{sec:concept_modality}, are considered as a set of instances and are aggregated via the \gls{mil}~\cite{abmil}.
As shown in Fig. 2, the \gls{mil} first projects the set of $f_i$ on to a feature space using a projector $H(\cdot)$, then employs a patch attention module $A^p(\cdot)$ to compute  attention $\alpha$ for each patch, given as,
\begin{equation}
\label{eq:patch_attention}
\tilde{f}_i = H(f_i); \hspace{0.3cm} \alpha_i = A^p(\tilde{f}_i); \hspace{0.3cm} i \in \{1,2,\dots,N\}
\end{equation}
\noindent Here, $A^p(\cdot)$ is a parameterized module with a softmax activation. Next, $\{\tilde{f}_i\}$ is attention-pooled using $\{\alpha_i\}$ to yield the \gls{wsi}-level deep-embedding $F_{wsi} \in \mathbb{R}^{D}$ as,
\begin{equation}
F_{wsi} =   \sum_{i=1}^{N} \alpha_i \cdot \tilde{f}_i
\end{equation}

\subsection{WSI-level Concept-Encoding Branch}
\label{sec:concept_encoding_branch}

The concept embedding is derived by aggregating concept prior from the  most salient patches. It involves identifying key patches, highlighting their salient concepts, and aggregating across the patches to compute the final concept embedding.

The top $K$ salient patches are identified using patch attention scores ${\alpha_i}$ (eq.~\ref{eq:patch_attention}), employing the differentiable \textbf{perturbed Top-$K$} operator as in~\cite{simil, zoommil} for optimal selection. The concept prior $M$ is truncated to these $K$ patches, rendering $\tilde{M} \in \mathbb{R}^{K \times C}$ for further processing.

We highlight the salient concepts by computing concept attention values from $\tilde{M}$. The transposed matrix $\tilde{M}^T$ is processed by an feature attention module, which uses \textbf{MLP-Mixer}~\cite{tolstikhin2021mlp} layers to contextualize spatial information across the $K$ patches and concept activation across the $C$ concepts. A gated attention network $G(\cdot)$ with sigmoid activation is then applied to each row of $\tilde{M}^T$, learning an attention score $\beta_j$ for each concept activation $C_j$, as in Eqn. \ref{eqn:betaconcept}.
$\beta_j$ are used to linearly transform $\tilde{M}$ into $\hat{M}$, as in Eqn. \ref{eqn:linearscale}, emphasizing salient concepts in a data-driven manner. To note, despite the non-linear operations to compute $\beta_j$, $\tilde{M}$ is only linearly scaled, ensuring the inherent interpretability of $\hat{M}$. Finally, the \gls{wsi}-level concept embedding $M_{wsi} \in \mathbb{R}^{C}$ is obtained by average pooling over $\hat{M}$, as in Eqn. \ref{eqn:conceptencoding}.
\begin{gather}
\beta_j = G\Big(\text{MLP-mixer}\big(\tilde{M}^T\big)\Big); \hspace{0.2cm} j \in \{1,...,C\} \label{eqn:betaconcept}\\
\hat{M}_{ij} = \beta_j \times \tilde{M}_{ij}; \hspace{0.2cm} i \in \{1,...,K\}, \hspace{0.1cm} j \in \{1,...,C\} \label{eqn:linearscale}\\
M_{wsi} = \frac{1}{K}\sum_{i=1}^{K}\hat{M}_{i} \label{eqn:conceptencoding}
\end{gather}
$M_{wsi}$ captures the activation of each concept at the \gls{wsi}-level, preserving its pathologist-friendly interpretability.

\subsection{GECKO: Pre-training and Inference}
\label{sec:concept_guided_pretrain}

\uline{\textbf{Pre-training:}} \gls{our} follows \gls{vcm} to align the \gls{wsi}-level embeddings $F_{wsi} \hspace{-0.1cm} \in \hspace{-0.1cm} \mathbb{R}^{D}$ and $M_{wsi} \hspace{-0.1cm} \in \hspace{-0.1cm} \mathbb{R}^{C}$ from the deep- and concept-encoding branches, respectively, in a shared latent space. To this end, we optimize the symmetric cross-modal CLIP~\cite{clip} contrastive learning loss $\mathcal{L}_{CL}$:

\begin{equation}
\resizebox{0.90\linewidth}{!}{$
\mathcal{L}_{CL}(a, b) = -\frac{1}{B} \sum_{i=1}^{B} \log \frac{\exp\big(\text{sim}(a^{i}, b^{i})/\tau\big)}{\sum_{j=1}^{B} \exp\big(\text{sim}(a^{i}, b^{j})/\tau\big)}
$}
\end{equation}
Here, $\text{sim}(\cdot, \cdot)$ denotes cosine similarity between embeddings, $B$ is batch size, and $\tau$ is a temperature hyperparameter. 
Note, a linear layer projects $F_{wsi}$ to match $M_{wsi}$ dimensions. The overall loss is computed as,
\begin{gather}
\mathcal{L} = \frac{1}{2}\Big(\mathcal{L}_{CL}(F_{wsi}, M_{wsi}) + \mathcal{L}_{CL}(M_{wsi}, F_{wsi})\Big) 
\end{gather}
Through this pretraining, the \gls{our}-pretrained dual-branch \gls{mil} is optimized to accurately identify discriminative \textit{patches} and \textit{concepts} within a \gls{wsi}, effectively contrasting them with other \glspl{wsi} in the batch. To enhance contrastive pretraining, we use false negative elimination~\cite{huynh2022boosting} with a keep ratio of \textbf{$r_{keep}$}. This process excludes a fraction (1 - \textbf{$r_{keep}$}) of highly similar \gls{wsi}-embeddings from the contrastive loss, preventing the comparison of similar \glspl{wsi}.

\noindent\uline{\textbf{Inference:}} After pretraining, the dual-branch MIL provides two embeddings: $F_{wsi} \in \mathbb{R}^{D}$ from the Deep-Encoding branch, and $M_{wsi} \in \mathbb{R}^{C}$ from the Concept-Encoding branch. $M_{wsi}$, being inherently interpretable, can be used for unsupervised prediction; and both $F_{wsi}$ and $M_{wsi}$ can be used for supervised-prediction.

\textbf{Unsupervised prediction:}
Let $I_l \subset I$ represent the indices of distinct concepts in $C$ corresponding to class $l \in L$ from the downstream task, where $I={0, 1, ..., C}$. The probability of a \gls{wsi} belonging to class $l$ is,
\begin{equation}
P(l) = \frac{\sum_{j \in I_l} M_{wsi,j}}{\sum_{k \in I} M_{wsi,k}}
\end{equation}
This pathologist-driven heuristic ensures that if the aggregated \gls{wsi}-level concepts for a class $l$ are predominantly activated, the probability $P(l)$ increases, similar to a pathological diagnosis. We refer to our model \gls{our}-Zero in this setting, as it requires no \gls{wsi}-level labels during training.

\textbf{Supervised prediction:}
Both $F_{wsi}$ and $M_{wsi}$ can be used for labeled classification via linear probing in both few-label and full-supervision setups. In our experiments, we evaluate them in both setups and also report results for an ``ensemble'' setup, where we average the predicted probabilities from using both embeddings.

\section{Experiments and Results}
\label{sec:experiments}

In this section, we provide the evaluation datasets and implementation details of the proposed \gls{our}-pretrained dual-branch \gls{mil} model across multiple \gls{wsi} classification tasks. We evaluate \gls{our} under unsupervised settings (with zero labels) as well as under supervised settings when few or all labels are utilized. The supervised setup also includes adapting \gls{our} to multi-modal scenarios (incorporating gene modality).

\begin{table*}[!ht]
\centering
\resizebox{\linewidth}{!}
{
\begin{tabular}{l c c c c c c}
\hline
\multirow{2}{*}{Methods} & Interpretable & LUAD vs. LUSC & EBV+MSI vs. Others & MSI vs. Others & EBV vs. Others & HER2 pos vs. neg vs. equi \\
 & (patch level-feature level) & (530 vs. 512) & (70 vs. 199) & (44 vs. 225) & (26 vs. 243) & (164 vs. 583 vs. 186) \\
\hline
MI-Zero~\cite{mizero}                
 & \cmark - \xmark 
 & \textbf{96.6} $\pm$ 1.4
 & 61.9 $\pm$ 6.1
 & 42.3 $\pm$ 5.3
 & 74.3 $\pm$ 11.9
 & 32.2 $\pm$ 1.3 \\
 
ConcepPath-Zero~\cite{conceppath}       
 & \xmark - \xmark 
 & 91.0 $\pm$ 3.3
 & 74.2 $\pm$ 7.2
 & 73.4 $\pm$ 7.9
 & 68.3 $\pm$ 11.2
 & 37.5 $\pm$ 1.1 \\

\gls{our}-Zero                 
 & \cmark - \cmark 
 & 95.0 $\pm$ 1.7
 & \textbf{83.4} $\pm$ 4.9
 & \textbf{77.1} $\pm$ 11.4
 & \textbf{82.5} $\pm$ 6.3
 & \textbf{60.6} $\pm$ 2.4 \\

\hline
ConcepPath~\cite{conceppath} (supervised)
 & \xmark - \xmark
 & 98.0  $\pm$ 0.7
 & 84.0  $\pm$ 5.5
 & 85.0  $\pm$ 5.0
 & 90.1  $\pm$ 4.0
 & 78.4  $\pm$ 1.2
 \\
\hline
\end{tabular}
}
\caption{Unsupervised classification analysis on TCGA datasets. Table shows AUC from  methods using no labels for supervision across tasks. Last row provides upper bound using ConcepPath~\cite{conceppath} under full supervision setting.}
\label{tab:unsupervised}
\vspace{-0.3cm}

\end{table*}

\subsection{Datasets and Implementation Details}
\label{sec:dataset}

\textbf{Datasets:} 
We evaluate our framework on three public datasets: TCGA-Lung, TCGA-STAD, and TCGA-BRCA, using the data splits from ConcepPath~\cite{conceppath}. Class details and \gls{wsi} distributions are provided in Supplementary Table \ref{tab:datasets}. In TCGA-STAD, we define three binary classification tasks: EBV vs. Others, MSI vs. Others, and EBV + MSI vs. Others. TCGA-Lung and TCGA-BRCA are used for binary and ternary classifications, respectively. To evaluate generalization on out-of-domain data, we use TCGA-Lung pretrained models for downstream analysis on the CPTAC-Lung dataset. All results are reported for 5-fold cross (mean, standard deviation) on TCGA datasets, with same splits provided by ConcepPath~\cite{conceppath}. Note that, for pretraining \gls{our} and baseline methods Intra and TANGLE, we only use training split for pretraining and conduct this pretraining separately for each fold to avoid any data leakage between train and test split.

\textbf{Defining Concepts:} 
For each task, we use a \gls{llm} with the concept generation technique from ConcepPath~\cite{conceppath} to produce concepts per class. In addition, we prompt the \gls{llm} to identify the 10 most visually distinct concepts per class from the generated concepts, producing $C=20$ for subtyping in TCGA-Lung and the three tasks in TCGA-STAD, and $C=30$ for the three-class TCGA-BRCA. A few example concepts in Lung cancer are provided in Fig.~\ref{fig:gecko_overview}, and the full list of concepts for all datasets is available in Supplementary Tables~\ref{tab:luad_vs_lusc_concepts}-~\ref{tab:her2_concepts}.

\textbf{Patch and Feature Extraction:} 
We use the vision and text encoders from the CONCH~\cite{conch} model to extract deep features $F$ and the concept prior $M$. Patches of size $448 \times 448$ pixels are extracted at $20\times$ magnification (0.5$\mu$m/px). Note that, unless specified otherwise, the CONCH model is used for deep feature extraction in our framework as well as in the baselines. We also perform ablation studies with different patch feature extractors for the deep encoding branch.

\textbf{\gls{mil} Setting:} 
For the deep-encoding branch, we use ABMIL~\cite{abmil} with the architecture from TANGLE~\cite{jaume2024transcriptomics}, featuring a 2-layer MLP projector $H(\cdot)$ with 512 hidden units and a gated-attention network $A^p(\cdot)$, also a 2-layer MLP with 512 hidden units and Sigmoid and Tanh activations. The concept-encoding branch employs 4 layers of MLP-Mixer and a gated-attention network $A^f(\cdot)$ with the same configuration, inspired by SI-MIL~\cite{simil}. We set $K=10$ by default. Since the concept-encoding branch input is $\tilde{M} \in \mathbb{R}^{K\times C}$, with $K=10$ and $C\in\{20, 30\}$, it is much more lightweight than the deep-encoding branch, which uses a 512-dimensional input from the CONCH-extracted features. Further implementation details can be found in Supplementary~\ref{implementation_details_additional}.

\subsection{Unsupervised Evaluation Setting}
\label{sec:unsupervised_setting}

\textbf{Baselines:} We benchmark our unsupervised predictive performance, \ie with \textbf{zero} \gls{wsi}-level annotations, against MI-zero~\cite{mizero} and ConcepPath-Zero, an adapted version of ConcepPath~\cite{conceppath}. To enable unsupervised prediction, we remove the learnable data-driven prompts and the trainable slide adapter from ConcepPath

\noindent \textbf{Results:} As observed in Table~\ref{tab:unsupervised}, \gls{our}-Zero outperforms MI-Zero by a significant 10--30\% margin across different tasks, except for the lung subtyping task where performance is comparable. Unlike MI-Zero, which directly uses slide-level classification prompts for cosine-similarity with patch-level features followed by pooling aggregation, ConcepPath-Zero decomposes slide-level prompts into concept prompts, often visible at patch-level, and then aggregates based on the activated concepts. This strategy yields considerable improvements for ConcepPath-Zero over MI-Zero, where relying on activation of slide-level prompts in each patch can introduce noisy aggregation. Notably, \gls{our}-Zero outperforms ConcepPath-Zero across all tasks for its ability of identifying salient regions and concepts in a \gls{wsi} via contrastive pretraining and aggregating concepts across the salient patches. To highlight, as our predictions are from \gls{wsi}-level concept embeddings, the predictions can be directly interpreted by pathologists, and  if required, can be corrected via test-time interventions~\cite{koh2020concept}.

\begin{figure*}[!ht]
    \centering
    \includegraphics[width=1\linewidth]{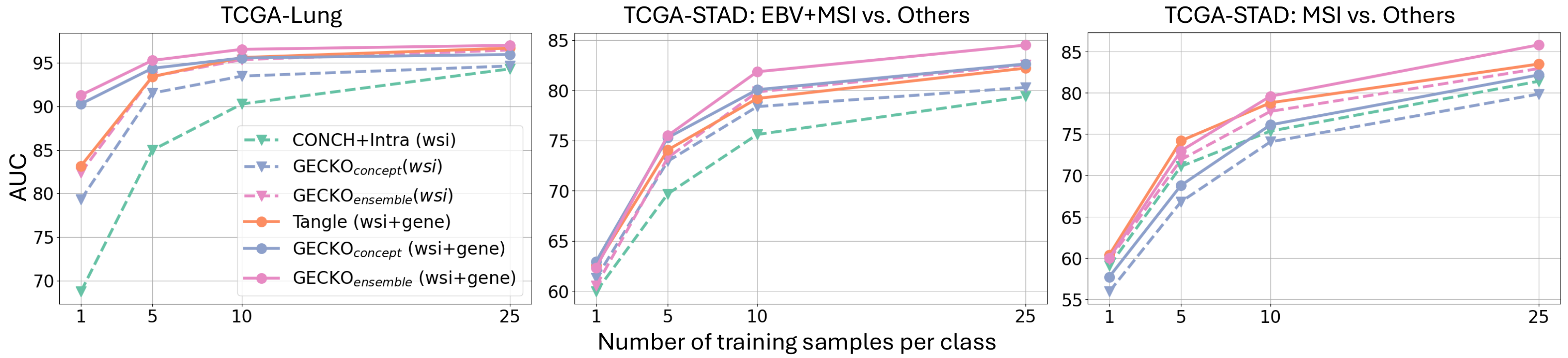}
    \caption{Few-labels (in-domain) classification analysis. The AUC results are obtained through linear probing. Dashed lines indicate pretraining using only \gls{wsi} data, while solid lines represent multimodal pretraining with additional transcriptomics data. CONCH is utilized for extracting deep features for image patches.}
    \label{fig:few_labels_plot}
\end{figure*}

\subsection{Supervised Evaluation Setting}
\label{sec:supervised_setting}

\textbf{Baselines:} Here we benchmark our method against two types of \gls{wsi}-level pretraining methods: (1) methods using only \glspl{wsi}, and (2) multimodal pretraining methods. Specifically, our baselines include unimodal pretraining method-- Intra established by~\cite{jaume2024transcriptomics,MADELEINE}, and  TANGLE~\cite{jaume2024transcriptomics} a multimodal pretraining method  incorporating gene data. 
We extract \gls{wsi}-level embeddings from the pretrained \gls{mil} aggregators and train linear classifiers using labels from the training dataset. We evaluate two scenarios: (1) few-$k$ labels per class, and (2) all training labels. The baseline methods are compared against ours (linear classifiers trained on \gls{wsi}-level deep embeddings (referred to as \gls{our}\textsubscript{deep}) and interpretable concept embeddings (referred to as \gls{our}\textsubscript{concept}) provided by \gls{our}-pretrained dual-branch \gls{mil} model). We also average the predicted probabilities from \gls{our}\textsubscript{deep} and \gls{our}\textsubscript{concept} to define \gls{our}\textsubscript{ensemble} predictions.

To address variability in the few-label setting, we perform 10 repetitions with shuffled samples from the training set for each $k$. For in-domain tasks, which involve pretraining on the train split, linear probing on few-$k$ samples per class from the training split, and testing on the test split across all TCGA datasets, this process is executed across all 5-folds (thus resulting in training 50 linear probing classifiers per dataset for each value of $k$ for our and baseline pretraining methods). We then report the average performance on the corresponding test sets over all repetitions and folds.
For out-of-domain generalization, we pretrain on the entire TCGA-Lung dataset, conduct linear probing on few-$k$ samples per class from the CPTAC-Lung dataset, and test on all remaining samples in CPTAC-Lung. We perform 10 repetitions by sampling different \glspl{wsi} for each value of $k$ and testing on all other \glspl{wsi} in the dataset. The mean and standard deviation of the results are reported in the Table~\ref{tab:few_labels_out_of_domain} (in Supplementary). We observe that when only WSI modality is available, \gls{our} significantly outperforms the baseline method Intra, and when gene modality is included, \gls{our} performs on par with TANGLE.

\begin{table*}[!ht]
\centering
\resizebox{\linewidth}{!}{
\begin{tabular}{c l c c  c c c c c}
\toprule
& \multirow{2}{*}{Methods} & \multirow{2}{*}{Embedding}  & Interpretable & LUAD vs. LUSC & EBV+MSI vs. Others & MSI vs. Others & EBV vs. Others & HER2 pos vs. neg vs. equi \\
& & & (patch level-feature level) & (530 vs. 512) & (70 vs. 199) & (44 vs. 225) & (26 vs. 243) & (164 vs. 583 vs. 186) \\
\midrule

\multirow{5}{*}{\rotatebox[origin=c]{90}{WSI only}} & *ConcepPath~\cite{conceppath}
& concept
 & \xmark - \xmark
 & \textbf{98.0 $\pm$ 0.7 }
 & 84.0 $\pm$ 5.5 
 & 85.0 $\pm$ 5.0 
 & \textbf{90.1 $\pm$ 4.0 }
 & \textbf{78.4 $\pm$ 1.2} \\

& Intra~\cite{jaume2024transcriptomics}
& deep
 & \cmark - \xmark
 & 97.5 $\pm$ 0.6 
 & 83.5 $\pm$ 8.3
 & 83.9 $\pm$ 7.0
 & 85.9 $\pm$ 5.2 
 & 74.2 $\pm$ 1.9 \\

\addlinespace[1pt]
  \cline{3-9}
\addlinespace[2pt]
  
& \multirow{3}{*}{\rotatebox[origin=c]{0}{\gls{our}}} 
& deep
 & \cmark - \xmark
 & 97.5 $\pm$ 0.3 
 & 85.3 $\pm$ 8.8 
 & 83.8 $\pm$ 4.8 
 & 84.3 $\pm$ 5.4 
 & 75.6 $\pm$ 1.5 \\

& 
& concept
 & \cmark - \cmark
 & 96.3 $\pm$ 1.2 
 & 82.8 $\pm$ 6.1 
 & 84.4 $\pm$ 10.9
 & 83.8 $\pm$ 6.1 
 & 75.9 $\pm$ 1.6 \\

&  
& ensemble
 & \cmark - \xmark
 & 97.6 $\pm$ 0.6
 & \textbf{86.4 $\pm$ 8.4}
 & \textbf{86.5 $\pm$ 7.8 }
 & 86.6 $\pm$ 6.0
 & 77.2 $\pm$ 1.6 \\

\midrule
\multirow{4}{*}{\rotatebox[origin=c]{90}{WSI + Gene}} 
& TANGLE~\cite{jaume2024transcriptomics}
& deep
 & \cmark - \xmark
 & \textbf{97.9 $\pm$ 0.5 }
 & 85.4 $\pm$ 8.0
 & 86.6 $\pm$ 5.0
 & 86.8 $\pm$ 2.7
 & 75.3 $\pm$ 1.3 \\

\addlinespace[1pt]
  \cline{3-9}
\addlinespace[2pt]
  
& \multirow{3}{*}{\rotatebox[origin=c]{0}{\gls{our}}} 
& deep
 & \cmark - \xmark
 & 97.8 $\pm$ 0.4
 & 86.1 $\pm$ 7.2 
 & 88.1 $\pm$ 4.3 
 & 86.6 $\pm$ 3.9
 & 76.4 $\pm$ 2.0 \\

&
& concept
 & \cmark - \cmark
 & 97.3 $\pm$ 0.8 
 & 85.6 $\pm$ 6.3 
 & 86.6 $\pm$ 6.5 
 & 82.7 $\pm$ 7.5
 & 76.3 $\pm$ 2.3 \\

&  
& ensemble
 & \cmark - \xmark
 & \textbf{97.9 $\pm$ 0.5} 
 & \textbf{87.1 $\pm$ 7.0} 
 & \textbf{89.4 $\pm$ 5.3 }
 & \textbf{87.4 $\pm$ 3.7 }
 & \textbf{78.4 $\pm$ 1.8} \\
\bottomrule
\end{tabular}
}
\caption{Full supervision (in-domain) classification analysis. Table compares AUC from various methods across multiple classification tasks. All results are with linear probing, except for *ConcepPath; In *ConcepPath, all parameters ($>$100K) are optimized with full supervision compared to just ($\sim$1K) parameters in linear probing. CONCH is utilized for extracting deep features.}
\label{tab:full_supervision}
\vspace{-0.3cm}

\end{table*}

\noindent \textbf{Few-Labels Setting:} As shown in Figure~\ref{fig:few_labels_plot}, in the unimodal setting with only \gls{wsi} data (indicated by dashed lines), linear probing on our interpretable \gls{wsi}-level concept embeddings (dim $C$) surpasses linear probing on Intra pretrained aggregator embeddings (dim $D>>C$) in several tasks, while also offering interpretable predictions. Moreover, \gls{our}\textsubscript{ensemble} consistently outperforms the Intra pretraining across all tasks.

In the multimodal setting (indicated by solid lines), \gls{our}\textsubscript{ensemble} uses TANGLE's gene encoding branch to pretrain with the gene modality alongside \glspl{wsi} and concept prior.  \gls{our}\textsubscript{ensemble} consistently outperforms the state-of-the-art TANGLE across all tasks. Interestingly, when \gls{our} is pretrained with the gene modality, the performance of linear probing with the interpretable concept embedding $M_{wsi}$ also improves significantly compared to pretraining with only \glspl{wsi}. Results for the EBV vs. Others and BRCA datasets in the few-label setting are provided in Supplementary Figure~\ref{fig:few_labels_plot_brca_ebv}.

\noindent \textbf{Full-Supervision Setting:} 
Table~\ref{tab:full_supervision} presents the results of using all training labels to supervise a linear classifier in a 5-fold cross-validation setting, comparing embeddings from our \gls{our}-pretrained model against the Intra and TANGLE baseline methods. Even with full supervision, our embeddings outperform Intra in the \gls{wsi}-only scenario and TANGLE when gene data is available. Remarkably, our dual-branch MIL aggregator, pretrained with \gls{our} without the additional gene modality, matches or exceeds TANGLE's performance, which relies on additional gene modality data. When the gene modality is included in \gls{our}, we observe further significant improvements across all tasks, except for the Lung dataset, where performance remains comparable.

Moreover, a linear classifier trained on our \gls{wsi}-level embeddings using only the \gls{wsi} modality outperforms ConcepPath in 2 out of 5 tasks, despite its lightweight nature. To note, all parameters of ConcepPath are fully optimized with label supervision, unlike ours, where we only train the linear classifier on slide-embeddings. When pretrained with the gene modality, our model significantly outperforms ConcepPath in 2 tasks and performs on par in 2 others. Although further fine-tuning with label supervision could enhance both our and baseline pretrained MIL backbones~\cite{MADELEINE,threads}, our primary goal is to establish a robust \gls{wsi} pretraining method that consistently outperforms both unimodal and multimodal pretraining approaches under few-labels and even full supervision conditions.

\subsection{Interpretability analysis}
\label{sec:Interpretability_analysis}

In Figure~\ref{fig:gecko_teaser}, we present the patch attention heatmap and the top four attended patches for a query \gls{wsi} using our \gls{our}-pretrained dual-branch \gls{mil}. Utilizing the interpretable \gls{wsi}-level concept embedding in an unsupervised setting, we highlight the two concepts with the highest and lowest aggregated activations and their activation strengths. Unlike previous MIL pretraining methods~\cite{jaume2024transcriptomics,lazard2023giga,MADELEINE} that could only provide patch attention maps, our approach uniquely identifies the pathology concepts driving predictions. In a user study, a pathologist reviewed \glspl{wsi} with saliency maps and ranked concepts, confirming that the most activated concepts matched the WSIs' predictions as LUSC/LUAD, consistent with the pathological prior. Additional examples are in Supplementary Sec.~\ref{interpretabiliy_analysis}.
Table~\ref{tab:quantitative_interpretability_transposed} presents a quantitative analysis of the accuracy of the salient concepts identified by our pretrained model in both unsupervised and supervised setting. 

\noindent \textbf{Unsupervised setting:} 
Using $M_{wsi}$, we extract the top-$j$ most activated concepts from a \gls{wsi}-level concept embedding and evaluate their accuracy based on overlap with the respective ground truth class concepts. The assessment is performed across 5-folds on the test split, reporting mean and standard deviation. Notably, the top-1 concept shows 81.4\% accuracy in Lung cancer and 54.0\% in the challenging MSI vs. Others task. Analysis across various $j$ reveals task complexity patterns: Lung cancer subtyping is relatively easy, making top-1 concept identification easy, but increasing $j$ mixes concepts across classes, reducing accuracy. Conversely, for MSI vs. Others, identifying the top-1 concept is challenging, but increasing $j$ improves accuracy by capturing correct class concepts. Importantly, concept selection excludes activation strengths, which explains why MSI vs. Others shows lower concept selection accuracy but higher prediction AUC in Table~\ref{tab:unsupervised}. This capability supports biomarker discovery in clinical settings with unlabeled \gls{wsi} data, enabling hypothesis testing by defining relevant concepts, pretraining a dual-branch MIL with \gls{our}, and analyzing salient concepts at the slide or class level using the interpretable $M_{wsi}$. \gls{our}'s interpretable-by-design architecture fundamentally sets it apart from existing interpretability methods.

\noindent \textbf{Supervised setting:} We multiply the interpretable concept embedding at the \gls{wsi}-level $M_{wsi}$, with the weights $w$ derived from the fully supervised trained linear classifier. When evaluating a test \gls{wsi}, the classifier's prediction determines the selection of concepts. 
If the prediction is class 0, we choose the concepts with the lowest values in $w \times M_{wsi}$. If the prediction is class 1, we choose the concepts with the highest values in $w \times M_{wsi}$. The reasoning for this is that a binary logistic regression classifier learns weights that push features related to class 0 towards lower values and features related to class 1 towards higher values, optimizing the sigmoid-based classification. Notably, we observe that in supervised setting, our method can identify top-1 concept with 99.9\% accuracy in Lung cancer and 83.3\% for the challenging MSI vs. Others task. The improved concept selection, compared to unsupervised setting, across different $j$ can be attributed to the strong label supervision.

\begin{table}[!t]
\centering
\resizebox{0.8\columnwidth}{!}{%
\begin{tabular}{c l c c}
\toprule
 & & Unsupervised & Fully-Supervised \\
\midrule

 LUAD & \(j=1\) & 81.4 \(\pm\) 2.5 & 99.9 \(\pm\) 0.2 \\
 vs. & \(j=3\) & 75.8 \(\pm\) 2.3 & 98.1 \(\pm\) 1.4 \\
LUSC & \(j=5\) & 71.6 \(\pm\) 1.2 & 95.1 \(\pm\) 3.5 \\
\midrule

MSI& \(j=1\) & 54.0 \(\pm\) 1.2 & 83.3 \(\pm\) 16.9 \\
vs. & \(j=3\) & 61.1 \(\pm\) 3.3 & 76.1 \(\pm\) 7.0 \\
Others& \(j=5\) & 61.0 \(\pm\) 3.9 & 80.5 \(\pm\) 3.3 \\

\bottomrule
\end{tabular}%
}
\caption{Accuracy of salient concepts identified by our pretrained model. Here, \(j\) denotes the correctness of the top-\(j\) concepts, determined by their alignment with the concepts associated with the ground-truth class.}
\label{tab:quantitative_interpretability_transposed}
\vspace{-0.3cm}

\end{table}

\subsection{Ablations}
\label{sec:slide_classification_performance}

\noindent \textbf{Comparison with additional \gls{wsi}-level encoding methods:} Thus far in this study, we have extensively compared \gls{our} against Intra in unimodal setting and TANGLE when gene data is available, consistently using the CONCH feature extractor. Table~\ref{tab:additional_slide_encoders} further compares \gls{our} with PANTHER~\cite{panther} and TITAN~\cite{titan}. Since TITAN utilizes the CONCH v1.5 encoder, we also base PANTHER on CONCH v1.5 embeddings and use CONCH v1.5 to extract deep features for \gls{our}. In addition, we evaluate \gls{wsi}-level embeddings from the unimodal Giga-SSL~\cite{lazard2023democratizing,lazard2023giga}, training a linear classifier. Giga-SSL provides \gls{wsi}-level embeddings based on pretraining on three different patch-level foundation models, Phikon~\cite{phikon}, Gigapath~\cite{gigapath}, and H-Optimus-0~\cite{hoptimus0}. Our method significantly outperforms PANTHER and Giga-SSL in \gls{wsi}-only settings across both datasets, with a notably larger performance gap at lower $k$ values.

\begin{table}[!t]
\centering
\resizebox{\columnwidth}{!}{%
\begin{tabular}{c l c  c c c c}
\toprule
& \multirow{2}{*}{Methods}  & \multirow{2}{*}{Embedding}    & \multicolumn{2}{c}{LUAD vs. LUSC}  & \multicolumn{2}{c}{EBV+MSI vs. Others} \\
&   &  & $k=10$  & $k=25$  & $k=10$ & $k=25$ \\

\midrule

  \multirow{6}{*}{\rotatebox[origin=c]{0}{WSI only}}  & PANTHER  & deep  &  91.2 $\pm$ 0.7  & 95.2 $\pm$  0.8  &  78.5 $\pm$ 2.8 &  84.0 $\pm$ 5.3 \\
  \addlinespace[1pt]
  \cline{3-7}
\addlinespace[2pt]

  & \multirow{2}{*}{Giga-SSL}  & Phikon  &  84.7 $\pm$ 1.7  & 89.5 $\pm$ 1.7  &  74.4 $\pm$ 5.6 &  79.4 $\pm$ 6.9 \\

  &  & Gigapath  &  92.4 $\pm$ 1.0  & 94.7 $\pm$ 1.0  &  78.0 $\pm$ 5.1 &  82.7 $\pm$ 6.2 \\

  &  & H-Optimus  &  92.8 $\pm$ 1.1  & 94.8 $\pm$ 1.0  &  77.5 $\pm$ 4.0 &  82.0 $\pm$ 5.4 \\

  \addlinespace[1pt]
  \cline{3-7}
\addlinespace[2pt]
  
 &   \multirow{2}{*}{\gls{our}} & concept    &  95.4 $\pm$ 0.7  & 95.6 $\pm$  0.7 &  75.9 $\pm$ 1.9 &  79.8 $\pm$ 3.8 \\
 &  & ensemble &  \textbf{96.4 $\pm$ 0.6}  &  
\textbf{96.7 $\pm$ 0.6}   & \textbf{ 82.1 $\pm$ 4.7} & \textbf{84.6 $\pm$ 5.5} \\

\midrule

 \multirow{3}{*}{\rotatebox[origin=c]{0}{Multi-modal}}  & TITAN & deep   &  \textbf{97.5 $\pm$ 0.9}  &  \textbf{97.7 $\pm$ 0.8}    &   
78.7 $\pm$ 3.6 & 84.7 $\pm$ 4.6 \\

  \addlinespace[1pt]
  \cline{3-7}
\addlinespace[2pt]
 &   \multirow{2}{*}{\gls{our}} & concept    &  96.2 $\pm$ 1.3  &  96.5 $\pm$ 1.0   &  79.3 $\pm$ 5.1 & 81.7 $\pm$ 5.7 \\
 &  & ensemble &  97.0 $\pm$ 1.1  & 97.2 $\pm$ 0.8   &  \textbf{84.4 $\pm$ 5.4} & \textbf{86.0 $\pm$ 5.8} \\
  
\bottomrule
\end{tabular}%
}
\caption{Comparison with additional \gls{wsi}-level encoding methods. All AUCs reported are with linear probing. CONCH v1.5 is used for extracting deep features.}
\label{tab:additional_slide_encoders}
\vspace{-0.3cm}

\end{table}

In the multimodal setting, TITAN is pretrained with closed-source pathology reports and synthetic captions. For a fair comparison, we evaluate \gls{our} using gene data and derived concept priors. While TITAN slightly surpasses our method by 0.5\% on the Lung dataset for $k={10,25}$, our approach significantly excels on the challenging EBV+MSI vs. Others dataset, with improvements of 6.7\% at $k=10$ and 1.3\% at $k=25$. Notably, TITAN is pretrained on over 100K paired multimodal samples, whereas \gls{our} uses only about 800 \glspl{wsi} for the Lung dataset and around 200 \glspl{wsi} for the EBV+MSI vs. Others STAD dataset. Furthermore, \gls{our} can incorporate the pathology reports used in TITAN's pretraining as an extra modality, potentially boosting performance further.

\noindent \textbf{Additional ablations.} In Supp. sec~\ref{additional_ablations}, we provide additional experiments to study (1) choice of \gls{wsi} Concept-Encoding branch architecture, and (2) the effect of the keep ratio (\textbf{$r_{keep}$}) in false negative elimination.

\section{Conclusion}
\label{sec:conclusion}

We introduce \gls{our}, a novel pretraining framework that learns robust \gls{wsi}-level representations without the need for auxiliary data modalities, while also delivering inherently interpretable \gls{wsi}-level concept embeddings. Additionally, \gls{our} effectively incorporates additional modalities when available, outperforming existing unimodal and multimodal methods and achieving consistent performance enhancements. In future research, we aim to explore pan-cancer pretraining with an expanded concept bank that encompasses a wide range of cancer sites and subtypes, setting the stage for a foundational slide-level \gls{our} capable of interpretable pan-cancer zero-shot prediction. Furthermore, our \gls{wsi}-level concept embedding could be utilized in generative models~\cite{yellapragada2024pathldm,graikos2024learned}, offering greater control in generating gigapixel images.

\section{Acknowledgment}
\label{sec:ack}

This research was partially supported by the National Institutes of Health (NIH) grants 1R01CA297843-01 and 3R21CA258493-02S1. The content is solely the responsibility of the authors and does not necessarily represent the official views of the NIH.

{
    \small
    \bibliographystyle{ieeenat_fullname}
    \bibliography{main}

\begin{thebibliography}{43}
\providecommand{\natexlab}[1]{#1}
\providecommand{\url}[1]{\texttt{#1}}
\expandafter\ifx\csname urlstyle\endcsname\relax
  \providecommand{\doi}[1]{doi: #1}\else
  \providecommand{\doi}{doi: \begingroup \urlstyle{rm}\Url}\fi

\bibitem[Barbiero et~al.(2023)Barbiero, Ciravegna, Giannini, Zarlenga, Magister, Tonda, Li{\'o}, Precioso, Jamnik, and Marra]{barbiero2023interpretable}
Pietro Barbiero, Gabriele Ciravegna, Francesco Giannini, Mateo~Espinosa Zarlenga, Lucie~Charlotte Magister, Alberto Tonda, Pietro Li{\'o}, Frederic Precioso, Mateja Jamnik, and Giuseppe Marra.
\newblock Interpretable neural-symbolic concept reasoning.
\newblock In \emph{International Conference on Machine Learning}, pages 1801--1825. PMLR, 2023.

\bibitem[Caron et~al.(2021)Caron, Touvron, Misra, J{\'e}gou, Mairal, Bojanowski, and Joulin]{dino}
Mathilde Caron, Hugo Touvron, Ishan Misra, Herv{\'e} J{\'e}gou, Julien Mairal, Piotr Bojanowski, and Armand Joulin.
\newblock Emerging properties in self-supervised vision transformers.
\newblock In \emph{Proceedings of the IEEE/CVF international conference on computer vision}, pages 9650--9660, 2021.

\bibitem[Chen et~al.(2022)Chen, Chen, Li, Chen, Trister, Krishnan, and Mahmood]{hipt}
Richard~J Chen, Chengkuan Chen, Yicong Li, Tiffany~Y Chen, Andrew~D Trister, Rahul~G Krishnan, and Faisal Mahmood.
\newblock Scaling vision transformers to gigapixel images via hierarchical self-supervised learning.
\newblock In \emph{Proceedings of the IEEE/CVF conference on computer vision and pattern recognition}, pages 16144--16155, 2022.

\bibitem[Chen et~al.(2024)Chen, Ding, Lu, Williamson, Jaume, Song, Chen, Zhang, Shao, Shaban, et~al.]{uni}
Richard~J Chen, Tong Ding, Ming~Y Lu, Drew~FK Williamson, Guillaume Jaume, Andrew~H Song, Bowen Chen, Andrew Zhang, Daniel Shao, Muhammad Shaban, et~al.
\newblock Towards a general-purpose foundation model for computational pathology.
\newblock \emph{Nature Medicine}, 30\penalty0 (3):\penalty0 850--862, 2024.

\bibitem[Ding et~al.(2024)Ding, Wagner, Song, Chen, Lu, Zhang, Vaidya, Jaume, Shaban, Kim, et~al.]{titan}
Tong Ding, Sophia~J Wagner, Andrew~H Song, Richard~J Chen, Ming~Y Lu, Andrew Zhang, Anurag~J Vaidya, Guillaume Jaume, Muhammad Shaban, Ahrong Kim, et~al.
\newblock Multimodal whole slide foundation model for pathology.
\newblock \emph{arXiv preprint arXiv:2411.19666}, 2024.

\bibitem[Espinosa~Zarlenga et~al.(2022)Espinosa~Zarlenga, Barbiero, Ciravegna, Marra, Giannini, Diligenti, Shams, Precioso, Melacci, Weller, et~al.]{cem}
Mateo Espinosa~Zarlenga, Pietro Barbiero, Gabriele Ciravegna, Giuseppe Marra, Francesco Giannini, Michelangelo Diligenti, Zohreh Shams, Frederic Precioso, Stefano Melacci, Adrian Weller, et~al.
\newblock Concept embedding models: Beyond the accuracy-explainability trade-off.
\newblock \emph{Advances in Neural Information Processing Systems}, 35:\penalty0 21400--21413, 2022.

\bibitem[Filiot et~al.(2023)Filiot, Ghermi, Olivier, Jacob, Fidon, Camara, Mac~Kain, Saillard, and Schiratti]{phikon}
Alexandre Filiot, Ridouane Ghermi, Antoine Olivier, Paul Jacob, Lucas Fidon, Axel Camara, Alice Mac~Kain, Charlie Saillard, and Jean-Baptiste Schiratti.
\newblock Scaling self-supervised learning for histopathology with masked image modeling.
\newblock \emph{medRxiv}, pages 2023--07, 2023.

\bibitem[Graikos et~al.(2024)Graikos, Yellapragada, Le, Kapse, Prasanna, Saltz, and Samaras]{graikos2024learned}
Alexandros Graikos, Srikar Yellapragada, Minh-Quan Le, Saarthak Kapse, Prateek Prasanna, Joel Saltz, and Dimitris Samaras.
\newblock Learned representation-guided diffusion models for large-image generation.
\newblock In \emph{Proceedings of the IEEE/CVF Conference on Computer Vision and Pattern Recognition}, pages 8532--8542, 2024.

\bibitem[Huang et~al.(2023)Huang, Bianchi, Yuksekgonul, Montine, and Zou]{huang2023visual}
Zhi Huang, Federico Bianchi, Mert Yuksekgonul, Thomas~J Montine, and James Zou.
\newblock A visual--language foundation model for pathology image analysis using medical twitter.
\newblock \emph{Nature medicine}, 29\penalty0 (9):\penalty0 2307--2316, 2023.

\bibitem[Huynh et~al.(2022)Huynh, Kornblith, Walter, Maire, and Khademi]{huynh2022boosting}
Tri Huynh, Simon Kornblith, Matthew~R Walter, Michael Maire, and Maryam Khademi.
\newblock Boosting contrastive self-supervised learning with false negative cancellation.
\newblock In \emph{Proceedings of the IEEE/CVF winter conference on applications of computer vision}, pages 2785--2795, 2022.

\bibitem[Ikezogwo et~al.(2023)Ikezogwo, Seyfioglu, Ghezloo, Geva, Sheikh~Mohammed, Anand, Krishna, and Shapiro]{ikezogwo2023quilt}
Wisdom Ikezogwo, Saygin Seyfioglu, Fatemeh Ghezloo, Dylan Geva, Fatwir Sheikh~Mohammed, Pavan~Kumar Anand, Ranjay Krishna, and Linda Shapiro.
\newblock Quilt-1m: One million image-text pairs for histopathology.
\newblock \emph{Advances in neural information processing systems}, 36:\penalty0 37995--38017, 2023.

\bibitem[Ilse et~al.(2018)Ilse, Tomczak, and Welling]{abmil}
Maximilian Ilse, Jakub Tomczak, and Max Welling.
\newblock Attention-based deep multiple instance learning.
\newblock In \emph{International conference on machine learning}, pages 2127--2136. PMLR, 2018.

\bibitem[Jaume et~al.(2021)Jaume, Pati, Bozorgtabar, Foncubierta, Anniciello, Feroce, Rau, Thiran, Gabrani, and Goksel]{jaume2021quantifying}
Guillaume Jaume, Pushpak Pati, Behzad Bozorgtabar, Antonio Foncubierta, Anna~Maria Anniciello, Florinda Feroce, Tilman Rau, Jean-Philippe Thiran, Maria Gabrani, and Orcun Goksel.
\newblock Quantifying explainers of graph neural networks in computational pathology.
\newblock In \emph{Proceedings of the IEEE/CVF conference on computer vision and pattern recognition}, pages 8106--8116, 2021.

\bibitem[Jaume et~al.(2024{\natexlab{a}})Jaume, Oldenburg, Vaidya, Chen, Williamson, Peeters, Song, and Mahmood]{jaume2024transcriptomics}
Guillaume Jaume, Lukas Oldenburg, Anurag Vaidya, Richard~J Chen, Drew~FK Williamson, Thomas Peeters, Andrew~H Song, and Faisal Mahmood.
\newblock Transcriptomics-guided slide representation learning in computational pathology.
\newblock In \emph{Proceedings of the IEEE/CVF Conference on Computer Vision and Pattern Recognition}, pages 9632--9644, 2024{\natexlab{a}}.

\bibitem[Jaume et~al.(2024{\natexlab{b}})Jaume, Vaidya, Zhang, H.~Song, J.~Chen, Sahai, Mo, Madrigal, Phi~Le, and Mahmood]{MADELEINE}
Guillaume Jaume, Anurag Vaidya, Andrew Zhang, Andrew H.~Song, Richard J.~Chen, Sharifa Sahai, Dandan Mo, Emilio Madrigal, Long Phi~Le, and Faisal Mahmood.
\newblock Multistain pretraining for slide representation learning in pathology.
\newblock In \emph{European Conference on Computer Vision}, pages 19--37. Springer, 2024{\natexlab{b}}.

\bibitem[Javed et~al.(2024)Javed, Mahmood, Ganapathi, Dharejo, Werghi, and Bennamoun]{javed2024cplip}
Sajid Javed, Arif Mahmood, Iyyakutti~Iyappan Ganapathi, Fayaz~Ali Dharejo, Naoufel Werghi, and Mohammed Bennamoun.
\newblock Cplip: zero-shot learning for histopathology with comprehensive vision-language alignment.
\newblock In \emph{Proceedings of the IEEE/CVF conference on computer vision and pattern recognition}, pages 11450--11459, 2024.

\bibitem[Kapse et~al.(2024)Kapse, Pati, Das, Zhang, Chen, Vakalopoulou, Saltz, Samaras, Gupta, and Prasanna]{simil}
Saarthak Kapse, Pushpak Pati, Srijan Das, Jingwei Zhang, Chao Chen, Maria Vakalopoulou, Joel Saltz, Dimitris Samaras, Rajarsi~R Gupta, and Prateek Prasanna.
\newblock Si-mil: Taming deep mil for self-interpretability in gigapixel histopathology.
\newblock In \emph{Proceedings of the IEEE/CVF Conference on Computer Vision and Pattern Recognition}, pages 11226--11237, 2024.

\bibitem[Koh et~al.(2020{\natexlab{a}})Koh, Nguyen, Tang, Mussmann, Pierson, Kim, and Liang]{cbm}
Pang~Wei Koh, Thao Nguyen, Yew~Siang Tang, Stephen Mussmann, Emma Pierson, Been Kim, and Percy Liang.
\newblock Concept bottleneck models.
\newblock In \emph{International conference on machine learning}, pages 5338--5348. PMLR, 2020{\natexlab{a}}.

\bibitem[Koh et~al.(2020{\natexlab{b}})Koh, Nguyen, Tang, Mussmann, Pierson, Kim, and Liang]{koh2020concept}
Pang~Wei Koh, Thao Nguyen, Yew~Siang Tang, Stephen Mussmann, Emma Pierson, Been Kim, and Percy Liang.
\newblock Concept bottleneck models.
\newblock In \emph{International conference on machine learning}, pages 5338--5348. PMLR, 2020{\natexlab{b}}.

\bibitem[Lazard et~al.(2023{\natexlab{a}})Lazard, Lerousseau, Decenci{\`e}re, and Walter]{lazard2023giga}
Tristan Lazard, Marvin Lerousseau, Etienne Decenci{\`e}re, and Thomas Walter.
\newblock Giga-ssl: Self-supervised learning for gigapixel images.
\newblock In \emph{Proceedings of the IEEE/CVF Conference on Computer Vision and Pattern Recognition}, pages 4305--4314, 2023{\natexlab{a}}.

\bibitem[Lazard et~al.(2023{\natexlab{b}})Lazard, Lerousseau, Gardrat, Vincent-Salomon, Stern, Rodrigues, Decenci{\`e}re, and Walter]{lazard2023democratizing}
Tristan Lazard, Marvin Lerousseau, Sophie Gardrat, Anne Vincent-Salomon, Marc-Henri Stern, Manuel Rodrigues, Etienne Decenci{\`e}re, and Thomas Walter.
\newblock Democratizing computational pathology: optimized whole slide image representations for the cancer genome atlas.
\newblock \emph{bioRxiv}, pages 2023--12, 2023{\natexlab{b}}.

\bibitem[Lu et~al.(2021)Lu, Williamson, Chen, Chen, Barbieri, and Mahmood]{clam}
Ming~Y Lu, Drew~FK Williamson, Tiffany~Y Chen, Richard~J Chen, Matteo Barbieri, and Faisal Mahmood.
\newblock Data-efficient and weakly supervised computational pathology on whole-slide images.
\newblock \emph{Nature biomedical engineering}, 5\penalty0 (6):\penalty0 555--570, 2021.

\bibitem[Lu et~al.(2023)Lu, Chen, Zhang, Williamson, Chen, Ding, Le, Chuang, and Mahmood]{mizero}
Ming~Y Lu, Bowen Chen, Andrew Zhang, Drew~FK Williamson, Richard~J Chen, Tong Ding, Long~Phi Le, Yung-Sung Chuang, and Faisal Mahmood.
\newblock Visual language pretrained multiple instance zero-shot transfer for histopathology images.
\newblock In \emph{Proceedings of the IEEE/CVF conference on computer vision and pattern recognition}, pages 19764--19775, 2023.

\bibitem[Lu et~al.(2024)Lu, Chen, Williamson, Chen, Liang, Ding, Jaume, Odintsov, Le, Gerber, et~al.]{conch}
Ming~Y Lu, Bowen Chen, Drew~FK Williamson, Richard~J Chen, Ivy Liang, Tong Ding, Guillaume Jaume, Igor Odintsov, Long~Phi Le, Georg Gerber, et~al.
\newblock A visual-language foundation model for computational pathology.
\newblock \emph{Nature Medicine}, 30\penalty0 (3):\penalty0 863--874, 2024.

\bibitem[Oquab et~al.(2023)Oquab, Darcet, Moutakanni, Vo, Szafraniec, Khalidov, Fernandez, Haziza, Massa, El-Nouby, et~al.]{dinov2}
Maxime Oquab, Timoth{\'e}e Darcet, Th{\'e}o Moutakanni, Huy Vo, Marc Szafraniec, Vasil Khalidov, Pierre Fernandez, Daniel Haziza, Francisco Massa, Alaaeldin El-Nouby, et~al.
\newblock Dinov2: Learning robust visual features without supervision.
\newblock \emph{arXiv preprint arXiv:2304.07193}, 2023.

\bibitem[Radford et~al.(2021)Radford, Kim, Hallacy, Ramesh, Goh, Agarwal, Sastry, Askell, Mishkin, Clark, et~al.]{clip}
Alec Radford, Jong~Wook Kim, Chris Hallacy, Aditya Ramesh, Gabriel Goh, Sandhini Agarwal, Girish Sastry, Amanda Askell, Pamela Mishkin, Jack Clark, et~al.
\newblock Learning transferable visual models from natural language supervision.
\newblock In \emph{International conference on machine learning}, pages 8748--8763. PmLR, 2021.

\bibitem[Rudin(2019)]{rudin2019stop}
Cynthia Rudin.
\newblock Stop explaining black box machine learning models for high stakes decisions and use interpretable models instead.
\newblock \emph{Nature machine intelligence}, 1\penalty0 (5):\penalty0 206--215, 2019.

\bibitem[Saillard et~al.(2024)Saillard, Jenatton, Llinares-López, Mariet, Cahané, Durand, and Vert]{hoptimus0}
Charlie Saillard, Rodolphe Jenatton, Felipe Llinares-López, Zelda Mariet, David Cahané, Eric Durand, and Jean-Philippe Vert.
\newblock H-optimus-0, 2024.

\bibitem[Seyfioglu et~al.(2024)Seyfioglu, Ikezogwo, Ghezloo, Krishna, and Shapiro]{seyfioglu2024quilt}
Mehmet~Saygin Seyfioglu, Wisdom~O Ikezogwo, Fatemeh Ghezloo, Ranjay Krishna, and Linda Shapiro.
\newblock Quilt-llava: Visual instruction tuning by extracting localized narratives from open-source histopathology videos.
\newblock In \emph{Proceedings of the IEEE/CVF Conference on Computer Vision and Pattern Recognition}, pages 13183--13192, 2024.

\bibitem[Song et~al.(2024)Song, Chen, Ding, Williamson, Jaume, and Mahmood]{panther}
Andrew~H Song, Richard~J Chen, Tong Ding, Drew~FK Williamson, Guillaume Jaume, and Faisal Mahmood.
\newblock Morphological prototyping for unsupervised slide representation learning in computational pathology.
\newblock In \emph{Proceedings of the IEEE/CVF Conference on Computer Vision and Pattern Recognition}, pages 11566--11578, 2024.

\bibitem[Sun et~al.(2025)Sun, Tessier, Meeuwsen, Grisi, van Midden, Litjens, and Baumgartner]{conceptmil}
Susu Sun, Leslie Tessier, Fr{\'e}d{\'e}rique Meeuwsen, Cl{\'e}ment Grisi, Dominique van Midden, Geert Litjens, and Christian~F Baumgartner.
\newblock Label-free concept based multiple instance learning for gigapixel histopathology.
\newblock \emph{arXiv preprint arXiv:2501.02922}, 2025.

\bibitem[Thandiackal et~al.(2022)Thandiackal, Chen, Pati, Jaume, Williamson, Gabrani, and Goksel]{zoommil}
Kevin Thandiackal, Boqi Chen, Pushpak Pati, Guillaume Jaume, Drew~FK Williamson, Maria Gabrani, and Orcun Goksel.
\newblock Differentiable zooming for multiple instance learning on whole-slide images.
\newblock In \emph{European Conference on Computer Vision}, pages 699--715. Springer, 2022.

\bibitem[Tolstikhin et~al.(2021)Tolstikhin, Houlsby, Kolesnikov, Beyer, Zhai, Unterthiner, Yung, Steiner, Keysers, Uszkoreit, et~al.]{tolstikhin2021mlp}
Ilya~O Tolstikhin, Neil Houlsby, Alexander Kolesnikov, Lucas Beyer, Xiaohua Zhai, Thomas Unterthiner, Jessica Yung, Andreas Steiner, Daniel Keysers, Jakob Uszkoreit, et~al.
\newblock Mlp-mixer: An all-mlp architecture for vision.
\newblock \emph{Advances in neural information processing systems}, 34:\penalty0 24261--24272, 2021.

\bibitem[Vaidya et~al.(2025)Vaidya, Zhang, Jaume, Song, Ding, Wagner, Lu, Doucet, Robertson, Almagro-Perez, et~al.]{threads}
Anurag Vaidya, Andrew Zhang, Guillaume Jaume, Andrew~H Song, Tong Ding, Sophia~J Wagner, Ming~Y Lu, Paul Doucet, Harry Robertson, Cristina Almagro-Perez, et~al.
\newblock Molecular-driven foundation model for oncologic pathology.
\newblock \emph{arXiv preprint arXiv:2501.16652}, 2025.

\bibitem[Vorontsov et~al.(2023)Vorontsov, Bozkurt, Casson, Shaikovski, Zelechowski, Liu, Severson, Zimmermann, Hall, Tenenholtz, et~al.]{virchow}
Eugene Vorontsov, Alican Bozkurt, Adam Casson, George Shaikovski, Michal Zelechowski, Siqi Liu, Kristen Severson, Eric Zimmermann, James Hall, Neil Tenenholtz, et~al.
\newblock Virchow: A million-slide digital pathology foundation model.
\newblock \emph{arXiv preprint arXiv:2309.07778}, 2023.

\bibitem[Watawana et~al.(2024)Watawana, Ranasinghe, Mahmood, Naseer, Khan, and Khan]{watawana2024hierarchical}
Hasindri Watawana, Kanchana Ranasinghe, Tariq Mahmood, Muzammal Naseer, Salman Khan, and Fahad~Shahbaz Khan.
\newblock Hierarchical text-to-vision self supervised alignment for improved histopathology representation learning.
\newblock In \emph{International Conference on Medical Image Computing and Computer-Assisted Intervention}, pages 167--177. Springer, 2024.

\bibitem[Xiang et~al.(2025)Xiang, Wang, Zhang, Xi, Eweje, Chen, Li, Bergstrom, Gopaulchan, Kim, et~al.]{xiang2025vision}
Jinxi Xiang, Xiyue Wang, Xiaoming Zhang, Yinghua Xi, Feyisope Eweje, Yijiang Chen, Yuchen Li, Colin Bergstrom, Matthew Gopaulchan, Ted Kim, et~al.
\newblock A vision--language foundation model for precision oncology.
\newblock \emph{Nature}, pages 1--10, 2025.

\bibitem[Xu et~al.(2024)Xu, Usuyama, Bagga, Zhang, Rao, Naumann, Wong, Gero, Gonz{\'a}lez, Gu, et~al.]{gigapath}
Hanwen Xu, Naoto Usuyama, Jaspreet Bagga, Sheng Zhang, Rajesh Rao, Tristan Naumann, Cliff Wong, Zelalem Gero, Javier Gonz{\'a}lez, Yu Gu, et~al.
\newblock A whole-slide foundation model for digital pathology from real-world data.
\newblock \emph{Nature}, 630\penalty0 (8015):\penalty0 181--188, 2024.

\bibitem[Yang et~al.(2023)Yang, Panagopoulou, Zhou, Jin, Callison-Burch, and Yatskar]{labo}
Yue Yang, Artemis Panagopoulou, Shenghao Zhou, Daniel Jin, Chris Callison-Burch, and Mark Yatskar.
\newblock Language in a bottle: Language model guided concept bottlenecks for interpretable image classification.
\newblock In \emph{Proceedings of the IEEE/CVF Conference on Computer Vision and Pattern Recognition}, pages 19187--19197, 2023.

\bibitem[Yellapragada et~al.(2024)Yellapragada, Graikos, Prasanna, Kurc, Saltz, and Samaras]{yellapragada2024pathldm}
Srikar Yellapragada, Alexandros Graikos, Prateek Prasanna, Tahsin Kurc, Joel Saltz, and Dimitris Samaras.
\newblock Pathldm: Text conditioned latent diffusion model for histopathology.
\newblock In \emph{Proceedings of the IEEE/CVF Winter Conference on Applications of Computer Vision}, pages 5182--5191, 2024.

\bibitem[Yu et~al.(2022)Yu, Wang, Vasudevan, Yeung, Seyedhosseini, and Wu]{coca}
Jiahui Yu, Zirui Wang, Vijay Vasudevan, Legg Yeung, Mojtaba Seyedhosseini, and Yonghui Wu.
\newblock Coca: Contrastive captioners are image-text foundation models.
\newblock \emph{arXiv preprint arXiv:2205.01917}, 2022.

\bibitem[Zhao et~al.(2024)Zhao, Guo, Fan, Jiang, Yeung, and Yu]{conceppath}
Weiqin Zhao, Ziyu Guo, Yinshuang Fan, Yuming Jiang, Maximus~CF Yeung, and Lequan Yu.
\newblock Aligning knowledge concepts to whole slide images for precise histopathology image analysis.
\newblock \emph{npj Digital Medicine}, 7\penalty0 (1):\penalty0 383, 2024.

\bibitem[Zhou et~al.(2024)Zhou, Zhang, Wu, Zhang, Xie, and Wang]{zhou2024knowledge}
Xiao Zhou, Xiaoman Zhang, Chaoyi Wu, Ya Zhang, Weidi Xie, and Yanfeng Wang.
\newblock Knowledge-enhanced visual-language pretraining for computational pathology.
\newblock In \emph{European Conference on Computer Vision}, pages 345--362. Springer, 2024.

\end{thebibliography}
}

\clearpage
\maketitlesupplementary

The supplementary presents the following materials: 
\begin{itemize}

   \item Generalizability Evaluation
    (Sec. \ref{generalizability_evaluation})

    \item Additional ablations 
    (Sec. \ref{additional_ablations})

    \item Additional implementation details
    (Sec. \ref{implementation_details_additional})

    \item  Additional few-labels supervised evaluation
    (Sec. \ref{supervised_evaluation_few_labels_additional})

    \item Interpretability analysis
    (Sec. \ref{interpretabiliy_analysis})

\end{itemize}

\section{Generalizability Evaluation}
\label{generalizability_evaluation}

Table~\ref{tab:few_labels_out_of_domain} presents the generalization ability of \gls{our} compared to Intra and TANGLE pretraining. When only the \gls{wsi} modality is available, both \gls{our}\textsubscript{deep} and \gls{our}\textsubscript{ensemble} significantly outperform at lower $k$ values and maintain superior performance at higher $k$ values. Additionally, when the gene modality is included, our performance matches that of TANGLE. Importantly, with gene modality in \gls{our} pretraining, the interpretable \gls{wsi}-level concept embedding consistently outperforms Intra pretraining, even on out-of-domain datasets. This demonstrates the potential to develop powerful aggregators that leverage multiple modalities for pretraining, offering inherently interpretable predictions that can build trust in clinical settings.

\begin{table}[!ht]
\centering
\resizebox{\columnwidth}{!}{%
\begin{tabular}{c l c c c c}
\toprule
  & Methods &  Embedding  & $k=5$ & $k=10$ & $k=25$ \\
\midrule
\multirow{4}{*}{\rotatebox[origin=c]{90}{WSI only}} 
  & Intra~\cite{jaume2024transcriptomics}  & deep           & 92.8 $\pm$ 2.1   & 95.0 $\pm$ 1.1   & 96.6 $\pm$ 0.7 \\
  
\addlinespace[1pt]

  \cline{3-6}
\addlinespace[2pt]  
  & \multirow{3}{*}{\rotatebox[origin=c]{0}{\gls{our}}} & deep    & 95.7 $\pm$ 1.2   & 96.2 $\pm$ 0.6   & 97.2 $\pm$ 0.7 \\
  &  & concept  & 93.0 $\pm$ 2.3   & 94.5 $\pm$ 1.8   & 95.7 $\pm$ 1.0 \\
  &  & ensemble & \textbf{96.3 $\pm$ 1.0}   & \textbf{96.9 $\pm$ 0.9}   & \textbf{97.3 $\pm$ 0.8} \\
\midrule
\multirow{4}{*}{\rotatebox[origin=c]{90}{WSI + Gene}} 
  & TANGLE~\cite{jaume2024transcriptomics}  & deep        & 97.0 $\pm$ 0.6   & 97.6 $\pm$ 0.6   & \textbf{98.3 $\pm$ 0.3} \\

\addlinespace[1pt]

  \cline{3-6}
\addlinespace[2pt]
  
  & \multirow{3}{*}{\rotatebox[origin=c]{0}{\gls{our}}} & deep    & \textbf{97.2 $\pm$ 0.7}   & \textbf{97.7 $\pm$ 0.6}   & \textbf{98.3 $\pm$ 0.3} \\
  & & concept  & 95.9 $\pm$ 1.2   & 95.7 $\pm$ 1.6   & 96.6 $\pm$ 0.7 \\
  & & ensemble & 97.0 $\pm$ 0.8   & 97.4 $\pm$ 0.6   & 97.9 $\pm$ 0.3 \\
\bottomrule
\end{tabular}%
}
\caption{Few Labels (out-of-domain) classification on binary CPTAC-Lung task. All AUCs are with linear probing. CONCH is used for extracting deep features.}
\label{tab:few_labels_out_of_domain}
\end{table}

\section{Additional ablations}
\label{additional_ablations}

\begin{figure}[!h]
    \centering
    \includegraphics[width=0.68\columnwidth]{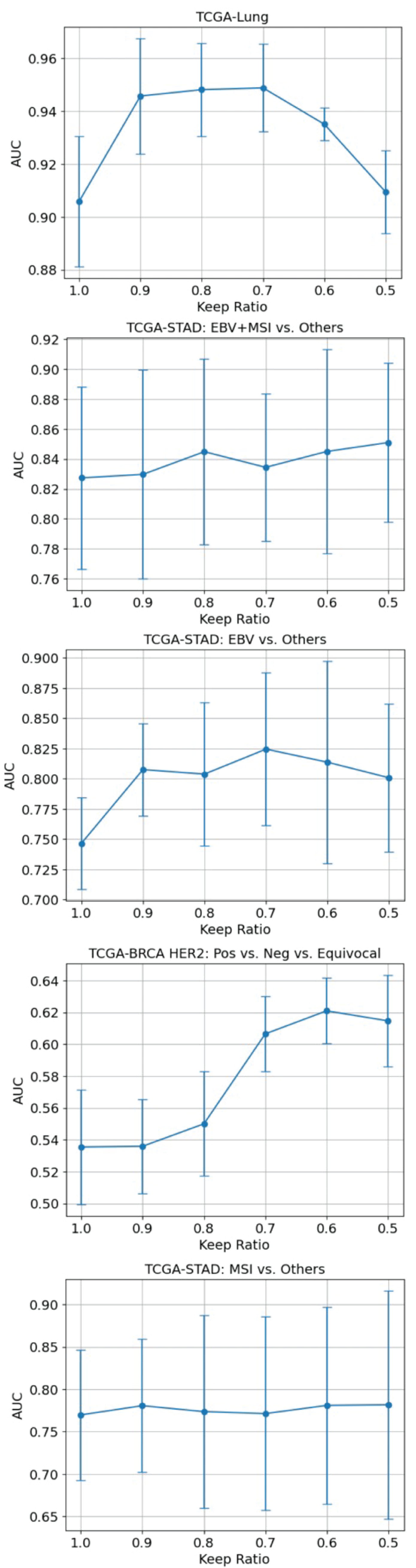}
    \caption{Effect of false negative elimination keep ratio (\textbf{$r_{keep}$}). All AUC values are reported in an unsupervised setting (5-fold cross validation) using our proposed heuristic. \textbf{$r_{keep}=0.7$} was found to work consistently well across all tasks.}
    \label{fig:Keep_ratio_ablation}
\end{figure}

\begin{enumerate}

\item \textbf{\gls{wsi} Concept-Encoding branch architecture:} By default, our dual-branch MIL uses the ABMIL~\cite{abmil} aggregator for the deep-encoding branch and a self-interpretable aggregator, inspired from SI-MIL~\cite{simil}, for the concept-encoding branch. For the ablation study, we replace the self-interpretable aggregator with an ABMIL in the concept-encoding branch that learns its own concept attention without reliance on the deep-encoding branch. As shown in Table~\ref{tab:additional_slide_mil_models}, our default dual-branch MIL (referred as \gls{our} in the Table) consistently outperforms the variant using a ABMIL for both branches (referred to as Dual-ABMIL). Note that we removed the projector $H(\cdot)$ in the ABMIL for the concept prior to enforce linear aggregation and thus preserve interpretability.

\begin{table}[!ht]
\centering
\resizebox{\columnwidth}{!}{%
\begin{tabular}{l c c c c c}
\toprule
\multirow{2}{*}{Methods}   & \multirow{2}{*}{Embedding}    & \multicolumn{2}{c}{LUAD vs. LUSC}  & \multicolumn{2}{c}{EBV+MSI vs. Others} \\

  &    & $k=10$  & $k=25$ & $k=10$ & $k=25$ \\

\midrule
  \multirow{2}{*}{Dual-ABMIL~\cite{abmil}} & concept         &  88.2 $\pm$ 0.7 & 90.3 $\pm$ 0.9 &  72.3 $\pm$ 5.9 & 73.8 $\pm$ 6.0  \\
  & ensemble & 92.3 $\pm$ 0.7  & 94.6 $\pm$ 1.0 & 75.1 $\pm$ 6.0 & 78.0 $\pm$ 7.1 \\


  \addlinespace[1pt]
  \cline{2-6}
\addlinespace[2pt]

   \multirow{2}{*}{\gls{our}}  & concept      & 93.5 $\pm$ 1.3  & 94.6 $\pm$ 1.5 &   78.4 $\pm$ 3.8 & 80.3 $\pm$ 6.1 \\
& ensemble &  \textbf{95.3 $\pm$ 0.9}  & \textbf{96.5 $\pm$ 1.1}   &  \textbf{79.8 $\pm$ 4.8} & \textbf{82.5 $\pm$ 7.4} \\
\bottomrule
\end{tabular}%
}
\caption{WSI Concept-Encoding branch architecture. All AUC results reported with linear probing, and pretraining with WSI only. CONCH is used for extracting deep features.}
\label{tab:additional_slide_mil_models}
\end{table}

\item   \textbf{Effect of false negative elimination with keep ratio (\textbf{$r_{keep}$}):} In Figure~\ref{fig:Keep_ratio_ablation}, we demonstrate the effect of \textbf{$r_{keep}$} for false negative elimination~\cite{huynh2022boosting} in contrastive pretraining across all five TCGA tasks. We report the performance of \gls{our}-Zero in an unsupervised 5-fold cross validation setting. We observed that the default contrastive pretraining with \textbf{$r_{keep}=1$} consistently results in poor performance. We attribute this to the fact that \gls{our} performs contrastive learning in $C$-dimensional embedding space, which is significantly smaller than a typical embedding size (256 or higher); thus, potentially contrasting \glspl{wsi} with similar concept activations and introducing noise. Recall that, we project the \gls{wsi}-level deep embedding to match the dimension of the \gls{wsi}-level concept embedding before alignment. Empirically, \textbf{$r_{keep}=0.7$} consistently performed well across tasks, thus we fix \textbf{$r_{keep}$} as 0.7 for our experiments.

\end{enumerate}

\section{Implementation details}
\label{implementation_details_additional}

\noindent \textbf{Pretraining setting.} We pretrained our dual-branch MIL using \gls{our} for 50 epochs for all tasks with a learning rate of 1e-4. A warmup is applied for 5 epochs, increasing the learning rate from 1e-8 to 1e-4, followed by a cosine scheduler that decays the rate to 1e-8, consistent with TANGLE~\cite{jaume2024transcriptomics}. The same settings were used to train TANGLE and Intra for all comparisons, with a batch size of 64 for all pretraining methods.
\\

\noindent \textbf{Linear probing setting.} For training the linear classifier across all methods, we use the same configuration as above. Specifically, we train LogisticRegression classifier from \texttt{sklearn} with default parameters and set the number of iterations to 10,000.
\\

\noindent \textbf{Gene modality setting.} For gene expression data, we adopt the curation method in~\cite{jaume2024transcriptomics,threads}, resulting in 4,848 gene expressions per case across all datasets. To integrate the gene modality into \gls{our}, we employ the same MLP-based architecture as in TANGLE. We perform K-way contrastive alignment by aligning each pair of modalities. To contrast with the concept prior, we use a projection head prior to alignment on the deep- and the gene-encoding branches to match the output dimension $C$ from the concept-branch. We directly align the outputs from the gene- and deep-encoding branches without additional projection, following TANGLE's design. Consequently, we optimize three losses in this multimodal setting of \gls{our}, which we average without any hyperparameter tuning.

\section{Additional few-label setting evaluation}
\label{supervised_evaluation_few_labels_additional}

\begin{figure*}[!ht]
    \centering
    \includegraphics[width=1\linewidth]{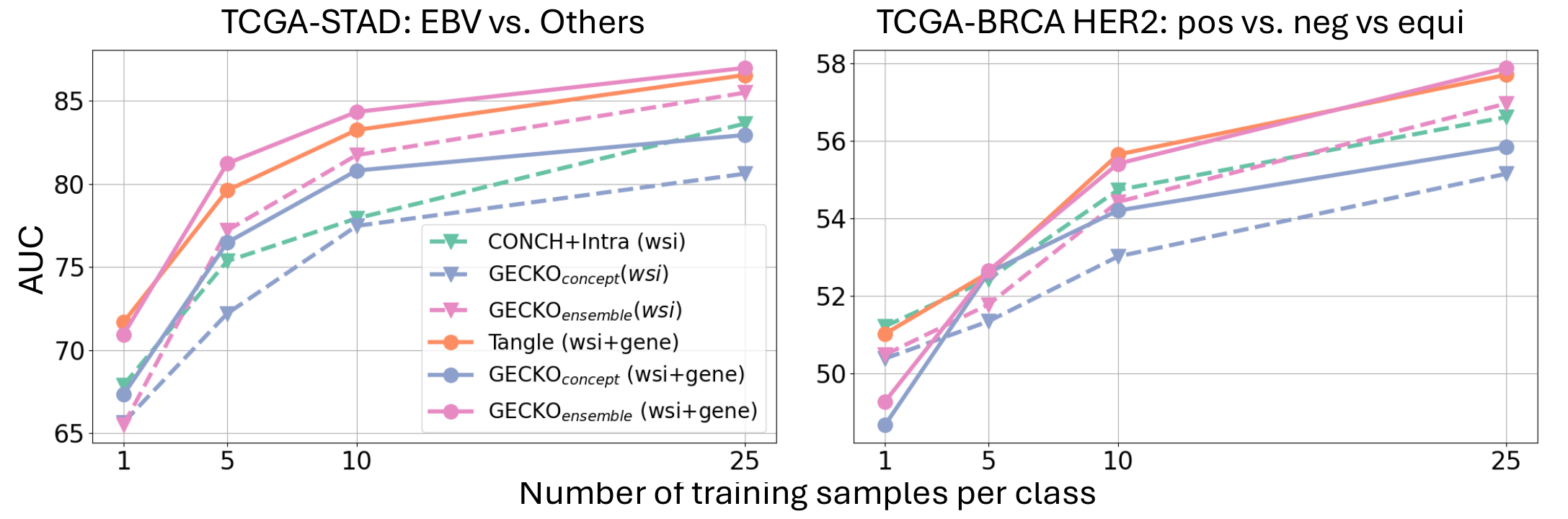}
    \caption{Few Labels (in domain) classification analysis. All AUC results are with linear probing. Dashed lines represent pretraining on WSI only, and solid lines represents multimodal pretraining with gene data. CONCH is used for extracting deep features.}
    \label{fig:few_labels_plot_brca_ebv}
\end{figure*}

In Figure~\ref{fig:few_labels_plot_brca_ebv}, we present the results of few-label supervised evaluation in the linear probing setting for the EBV vs. Others and BRCA datasets. In the unimodal setting with only \gls{wsi} data (indicated by dashed lines), our \gls{our}\textsubscript{ensemble} significantly outperforms the Intra pretraining on the EBV vs. Others task, while performing on par for HER2 prediction. In the multimodal setting, where gene data is available (indicated by solid lines), \gls{our}\textsubscript{ensemble} pretrained with the gene modality alongside \glspl{wsi} and our concept prior slightly outperforms TANGLE on the EBV vs. Others task, while achieving comparable performance on the HER2 prediction task.

\section{Interpretability analysis}
\label{interpretabiliy_analysis}

\begin{figure*}[!ht]
    \centering
    \includegraphics[width=1\linewidth]{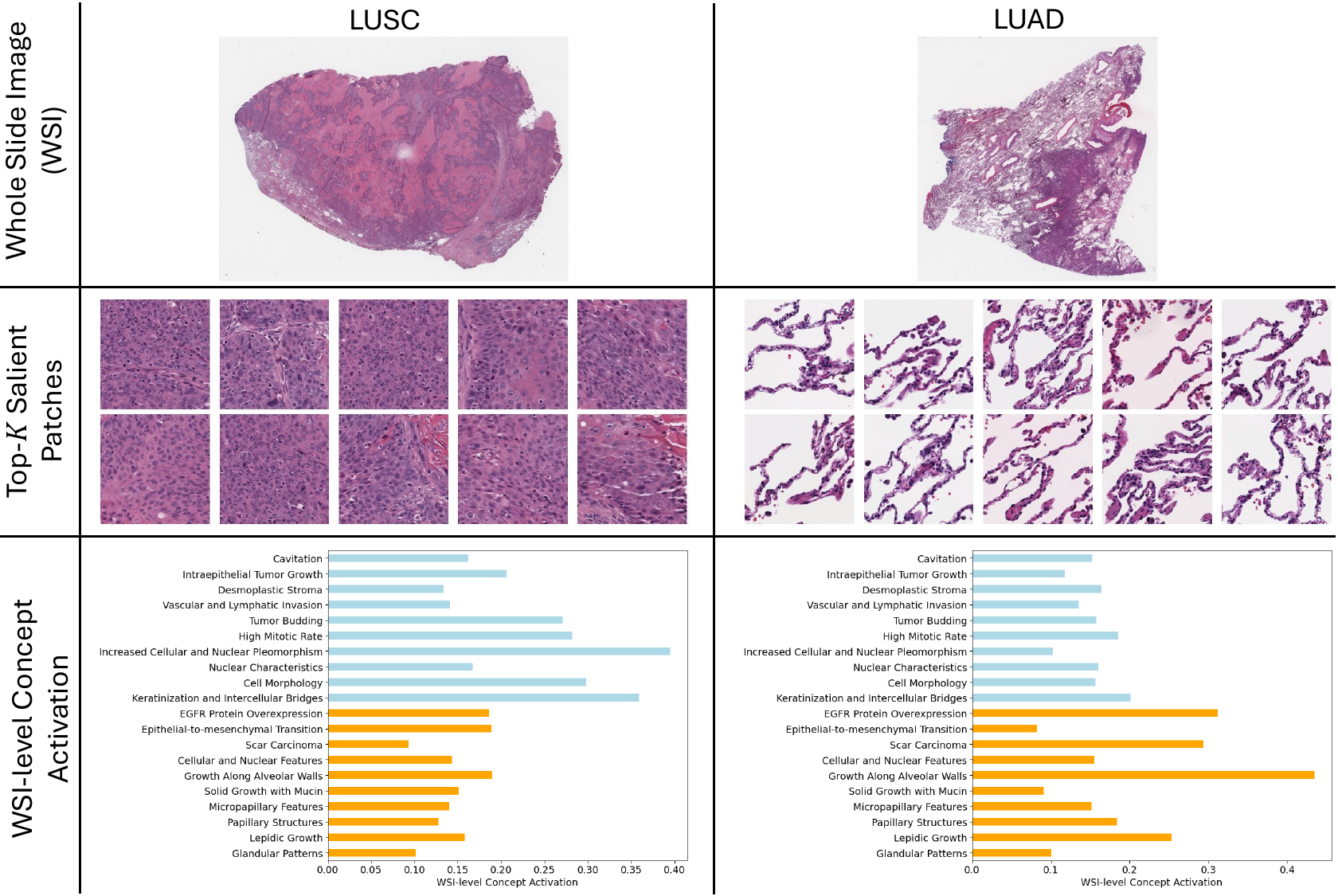}
    \caption{TCGA-Lung: \textcolor{blue!30}{LUSC} vs. \textcolor{orange}{LUAD}. Row 1 shows sample WSIs from LUSC and LUSC subtypes in TCGA-Lung. Row 2 shows the Top-$K$ patches selected by our \gls{our} pretrained model. Row 3 illustrates the WSI-level aggregated concept activation (from interpretable concept embedding).}
    \label{fig:GECKO_interpretability_supp_lusc_luad}
\end{figure*}

\begin{figure*}[!ht]
    \centering
    \includegraphics[width=1\linewidth]{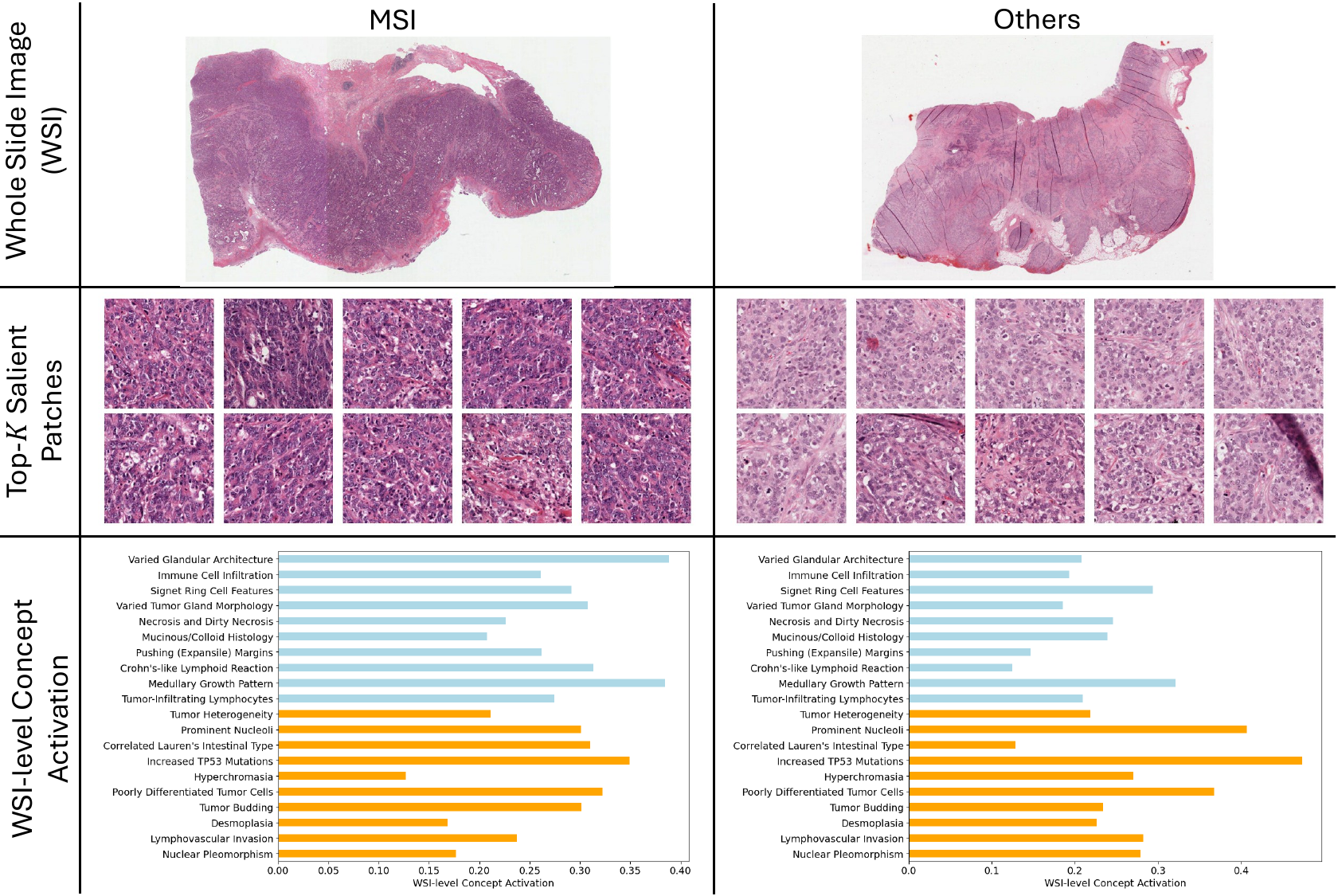}
    \caption{TCGA-STAD: \textcolor{blue!30}{MSI} vs. \textcolor{orange}{Others}. Row 1 shows sample WSIs from MSI and Others class in TCGA-STAD. Row 2 shows the Top-$K$ patches selected by our \gls{our} pretrained model. Row 3 illustrates the WSI-level aggregated concept activation (from interpretable concept embedding).}
    \label{fig:GECKO_interpretability_supp_msi_vs_others}
\end{figure*}

In Fig.~\ref{fig:GECKO_interpretability_supp_lusc_luad} and~\ref{fig:GECKO_interpretability_supp_msi_vs_others}, we illustrate the Top-$K$ salient patches and the WSI-level concept activations produced by our \gls{our}-pretrained model for TCGA-Lung and TCGA-STAD in an unsupervised setting. In the WSI-level concept activation bar plots, we quantitatively demonstrate that for a WSI belonging to a particular class, our model not only identifies the important patches but also provides the WSI-level activation for each concept through its interpretable concept embedding $M_{wsi}$. Notably, the concepts with the highest activations align with those that are most relevant to the corresponding class, evaluated by a pathologist. In Table~\ref{tab:luad_vs_lusc_concepts}--\ref{tab:her2_concepts}, we provide the concepts for each task along with their detailed descriptions, that were used as input to the text encoder of the CONCH model, in line with ConcepPath~\cite{conceppath}.

\begin{table*}[!ht]
 \centering
 \resizebox{0.8\textwidth}{!}{%
 \begin{tabular}{lcl}
 \toprule
 \textbf{Dataset} & \textbf{\#\glspl{wsi}} & \textbf{Class name (\#\glspl{wsi})} \\
 \midrule
 \multirow{2}{*}{TCGA-Lung} & \multirow{2}{*}{1042} & LUAD: Lung adenocarcinoma (530) \\
 & & LUSC: Lung squamous cell carcinoma (512) \\
 \cmidrule(lr){1-3}
 \multirow{3}{*}{TCGA-BRCA} & \multirow{3}{*}{933} & HER2-positive (164) \\
 & & Equivocal (186) \\
 & & HER2-Negative (583) \\
 \cmidrule(lr){1-3}
 \multirow{3}{*}{TCGA-STAD} & \multirow{3}{*}{268} & EBV: Epstein-Barr virus (26) \\
 & & MSI: Microsatellite Instability (44) \\
 & & GS:Genomically Stable/CIN: Chromosomally Instable (199) \\
 \cmidrule(lr){1-3}
 \multirow{2}{*}{CPTAC-Lung} & \multirow{2}{*}{1091} & LUAD: Lung adenocarcinoma (578) \\
 & & LUSC: Lung squamous cell carcinoma (513) \\
 \bottomrule
 \end{tabular}
 }
 \caption{Datasets (with class distribution) used for evaluation .}
 \label{tab:datasets}
\end{table*}

\begin{table*}[ht]
\centering
\resizebox{\linewidth}{!}{
\begin{tabular}{l l l}
\toprule
\textbf{Type} & \textbf{Concept} & \textbf{Description} \\ 
\midrule
\multirow{10}{*}{LUAD} 
  & Glandular Patterns & Gland-like structures; tubular; acinar; papillary formations; lined by atypical cells; mucin production; \\
\addlinespace[2pt]
\cline{2-3}
\addlinespace[2pt]
  & Lepidic Growth & Alveolar growth pattern; non-invasive; early adenocarcinomas; minimally invasive adenocarcinomas; \\
\addlinespace[2pt]
\cline{2-3}
\addlinespace[2pt]
  & Papillary Structures & Papillary architecture; fibrovascular cores; malignant cells lining; mucin content; \\
\addlinespace[2pt]
\cline{2-3}
\addlinespace[2pt]
  & Micropapillary Features & Micropapillary pattern; small cell clusters; no fibrovascular core; clear spaces from tissue processing; \\
\addlinespace[2pt]
\cline{2-3}
\addlinespace[2pt]
  & Solid Growth with Mucin & Solid growth pattern with mucin; mucicarmine; periodic acid-Schiff stains usage; \\
\addlinespace[2pt]
\cline{2-3}
\addlinespace[2pt]
  & Growth Along Alveolar Walls & Lepidic growth pattern; tumor cells along alveolar walls; non-invasive; \\
\addlinespace[2pt]
\cline{2-3}
\addlinespace[2pt]
  & Cellular and Nuclear Features & Cell morphology variable; cuboidal to columnar shape; hobnail appearance; pleomorphic nuclei; prominent nucleoli; \\
\addlinespace[2pt]
\cline{2-3}
\addlinespace[2pt]
  & Scar Carcinoma & Association with lung scarring or fibrosis; possible misdiagnosis on imaging; requires biopsy; \\
\addlinespace[2pt]
\cline{2-3}
\addlinespace[2pt]
  & Epithelial-to-mesenchymal Transition & E-cadherin staining decrease; mesenchymal markers increase; cytoplasmic/membranous staining; EMT at invasive front; \\
\addlinespace[2pt]
\cline{2-3}
\addlinespace[2pt]
  & EGFR Protein Overexpression & EGFR expression; membranous staining; possible cytoplasmic staining; cell membrane receptor; \\
\midrule
\multirow{10}{*}{LUSC} 
  & Keratinization and Intercellular Bridges & Squamous differentiation; keratin production; keratin pearls; intercellular bridges; \\
\addlinespace[2pt]
\cline{2-3}
\addlinespace[2pt]
  & Cell Morphology & Polygonal tumor cells; abundant eosinophilic cytoplasm; high keratin content; \\
\addlinespace[2pt]
\cline{2-3}
\addlinespace[2pt]
  & Nuclear Characteristics & Hyperchromatic nuclei; prominent nucleoli; variable pleomorphism; \\
\addlinespace[2pt]
\cline{2-3}
\addlinespace[2pt]
  & Increased Cellular and Nuclear Pleomorphism & Cellular and nuclear pleomorphism; increased variability; indicative of higher malignancy; IHC highlighted; \\
\addlinespace[2pt]
\cline{2-3}
\addlinespace[2pt]
  & High Mitotic Rate & High mitotic figure count; rapid cell proliferation; visualized by mitotic markers; \\
\addlinespace[2pt]
\cline{2-3}
\addlinespace[2pt]
  & Tumor Budding & Tumor budding presence; aggressive behavior indicator; cytokeratin stains highlight; \\
\addlinespace[2pt]
\cline{2-3}
\addlinespace[2pt]
  & Vascular and Lymphatic Invasion & Tumor cells in blood vessels or lymphatics; potential for metastasis; CD31 and podoplanin (D2-40) markers; \\
\addlinespace[2pt]
\cline{2-3}
\addlinespace[2pt]
  & Desmoplastic Stroma & Reactive stromal response; dense fibrous stroma surrounding tumor cells; \\
\addlinespace[2pt]
\cline{2-3}
\addlinespace[2pt]
  & Intraepithelial Tumor Growth & Intraepithelial growth; tumor spread within epithelial structures; \\
\addlinespace[2pt]
\cline{2-3}
\addlinespace[2pt]
  & Cavitation & Cavitation; central necrosis; more common in squamous cell carcinoma; visible on imaging; \\
\bottomrule
\end{tabular}
}
\caption{Pathology concepts for LUAD vs. LUSC}
\label{tab:luad_vs_lusc_concepts}

\end{table*}

\begin{table*}[ht]
\centering
\resizebox{\linewidth}{!}{
\begin{tabular}{l l l}
\toprule
\textbf{Type} & \textbf{Concept} & \textbf{Description} \\ 
\midrule

\multirow{15}{*}{EBV+MSI} 
  & Lymphoepithelioma-like Histology & \makecell[l]{EBV-positive; lymphoepithelioma-like carcinoma; large sheets; syncytial clusters; undifferentiated cells; \\ prominent lymphoid infiltration; no glandular formation; \\ non-keratinizing; vesicular nuclei; prominent nucleoli; desmoplastic reaction;} \\
\addlinespace[2pt]
\cline{2-3}
\addlinespace[2pt]
  & Syncytial trabecular pattern & Syncytial trabecular pattern; nested growth; cord-like structures; indistinct cell borders; interconnected net-like structure; \\
\addlinespace[2pt]
\cline{2-3}
\addlinespace[2pt]
  & Tumor Infiltrating Lymphocytes & Tumor-infiltrating lymphocytes; dispersed or clustered; infiltrating between cells or stromal; indicative of immune response; \\
\addlinespace[2pt]
\cline{2-3}
\addlinespace[2pt]
  & Intraepithelial Lymphocytosis & \makecell[l]{Intraepithelial lymphocytes; small, round; dense nuclei; disrupts architecture; associated with neoplastic epithelium; \\ stromal lymphoplasmacytic infiltration;} \\
\addlinespace[2pt]
\cline{2-3}
\addlinespace[2pt]
  & Stromal Lymphoplasmacytic Infiltration & \makecell[l]{Lymphocytes in stroma; plasma cells present; small cells with large nuclei; abundant basophilic cytoplasm; \\ interspersed infiltration; reactive changes; possible fibrosis or edema;} \\
\addlinespace[2pt]
\cline{2-3}
\addlinespace[2pt]
  & Medullary Growth Pattern & \makecell[l]{Carcinomas; colorectal; MSI-H status; high neoantigen load; poorly differentiated; syncytial growth; \\ abundant intraepithelial lymphocytes; dMMR tumors; solid sheets of cells;} \\
\addlinespace[2pt]
\cline{2-3}
\addlinespace[2pt]
  & Crohn's-like Lymphoid Reaction & Dense lymphoid aggregates; tumor margin; robust immune response; neoantigens; dMMR tumors; Crohn's-like reaction; \\
\addlinespace[2pt]
\cline{2-3}
\addlinespace[2pt]
  & Pushing (Expansile) Margins & \makecell[l]{Expansive growth pattern; pushing borders; high neoantigen levels; immune containment; dMMR tumors; \\ non-infiltrative margin; microsatellite stable (MSS) tumors contrast;} \\
\addlinespace[2pt]
\cline{2-3}
\addlinespace[2pt]
  & Pattern of Infiltration & \makecell[l]{Vigorous immune infiltrate; variable PD-L1 positive cell distribution; invasive tumor margins; tumor nests; \\ 'brisk' infiltration pattern; T cell band at tumor margin; 'non-brisk' infiltration pattern; \\ scattered T cells throughout tumor;} \\
\addlinespace[2pt]
\cline{2-3}
\addlinespace[2pt]
  & Immune Cell Infiltration & Significant number; lymphocytes; tumor tissue presence; \\

\midrule

\multirow{15}{*}{Others} 
  & Nuclear Pleomorphism & \makecell[l]{Variation in nuclear size and shape; nuclei size disparity; irregular nuclear shapes; \\ oval to highly irregular forms;} \\
\addlinespace[2pt]
\cline{2-3}
\addlinespace[2pt]
  & Hyperchromasia & Nuclei appear darker; excess DNA content; \\
\addlinespace[2pt]
\cline{2-3}
\addlinespace[2pt]
  & Irregular Nuclear Contours & Uneven nuclear borders; indented nuclear contours; \\
\addlinespace[2pt]
\cline{2-3}
\addlinespace[2pt]
  & Prominent Nucleoli & Prominent nucleoli; increased number of nucleoli; sign of heightened protein synthesis; rapid cell division indicator; \\
\addlinespace[2pt]
\cline{2-3}
\addlinespace[2pt]
  & Chromatin Clumping & Irregular chromatin clumping; patchy nuclear appearance; \\
\addlinespace[2pt]
\cline{2-3}
\addlinespace[2pt]
  & Multipolar spindles & \makecell[l]{Multipolar spindles; asymmetric nuclear division; uneven genetic material distribution; \\ cells with abnormal nuclear shapes and \\ sizes;} \\
\addlinespace[2pt]
\cline{2-3}
\addlinespace[2pt]
  & Lymphovascular Invasion & Tumor cells in lymphatic vessels; tumor cells in blood vessels; direct indication of metastasis; \\
\addlinespace[2pt]
\cline{2-3}
\addlinespace[2pt]
  & Tumor Budding & \makecell[l]{Small clusters of cancer cells at invasive front; individual cells at invasive front; \\ sign of aggressive tumor phenotype; correlated with metastasis;} \\
\addlinespace[2pt]
\cline{2-3}
\addlinespace[2pt]
  & Desmoplasia & Pronounced desmoplastic reaction; growth of fibrous tissue; connective tissue increase; association with aggressive tumors; \\
\addlinespace[2pt]
\cline{2-3}
\addlinespace[2pt]
  & Signet Ring Cells & \makecell[l]{Loss of E-cadherin function; CDH1 mutations; presence of signet ring cells; large vacuole in cells; \\ nucleus at periphery; signet ring-like appearance; indicative of poor prognosis;} \\

\bottomrule
\end{tabular}
}
\caption{Pathology concepts for EBV+MSI vs. Others}
\label{tab:ebvnmsi_vs_others_concepts}

\end{table*}

\begin{table*}[ht]
\centering
\resizebox{\linewidth}{!}{
\begin{tabular}{l l l}
\toprule
\textbf{Type} & \textbf{Concept} & \textbf{Description} \\ 
\midrule

\multirow{15}{*}{MSI} 
  & Tumor-Infiltrating Lymphocytes & \makecell[l]{High neoantigen load; immune cell infiltration; tumor tissue response; neoantigen presentation; \\ dMMR tumors; prominent lymphocytic \\ response; high TIL density; immune response to neoantigens;} \\
\addlinespace[2pt]
\cline{2-3}
\addlinespace[2pt]
  & Medullary Growth Pattern & \makecell[l]{Carcinomas; colorectal; MSI-H status; high neoantigen load; poorly differentiated; syncytial growth; \\ abundant intraepithelial lymphocytes; dMMR \\ tumors; solid sheets of cells;} \\
\addlinespace[2pt]
\cline{2-3}
\addlinespace[2pt]
  & Crohn's-like Lymphoid Reaction & Dense lymphoid aggregates; tumor margin; robust immune response; neoantigens; dMMR tumors; Crohn's-like reaction; \\
\addlinespace[2pt]
\cline{2-3}
\addlinespace[2pt]
  & Pushing (Expansile) Margins & \makecell[l]{Expansive growth pattern; pushing borders; high neoantigen levels; immune containment; dMMR tumors; \\ non-infiltrative margin; microsatellite \\ stable (MSS) tumors contrast;} \\
\addlinespace[2pt]
\cline{2-3}
\addlinespace[2pt]
  & Mucinous/Colloid Histology & Abundance; extracellular mucin production; MSI-H tumors; \\
\addlinespace[2pt]
\cline{2-3}
\addlinespace[2pt]
  & Necrosis and Dirty Necrosis & \makecell[l]{High neoantigen loads; necrosis; cytotoxic immune response; tumor necrosis; 'dirty necrosis'; \\ debris; nuclear dust; dMMR tumors \\ commonality;} \\
\addlinespace[2pt]
\cline{2-3}
\addlinespace[2pt]
  & Varied Tumor Gland Morphology & dMMR tumors; heterogeneous morphology; varied gland shapes; varied gland sizes; poor differentiation; \\
\addlinespace[2pt]
\cline{2-3}
\addlinespace[2pt]
  & Signet Ring Cell Features & Mucin-filled cells; peripheral nucleus; indicative of MSI-H; gastric cancer; \\
\addlinespace[2pt]
\cline{2-3}
\addlinespace[2pt]
  & Immune Cell Infiltration & Significant number; lymphocytes; tumor tissue presence; \\
\addlinespace[2pt]
\cline{2-3}
\addlinespace[2pt]
  & Varied Glandular Architecture & Disorganized structure; irregular gland formation; varied gland sizes; MSI-H tumors; \\
  \midrule

\multirow{15}{*}{Others} 
  & Nuclear Pleomorphism & \makecell[l]{Variation in nuclear size and shape; nuclei size disparity; irregular nuclear shapes; \\ oval to highly \\ irregular forms;} \\
\addlinespace[2pt]
\cline{2-3}
\addlinespace[2pt]
  & Lymphovascular Invasion & Tumor cells in lymphatic vessels; tumor cells in blood vessels; direct indication of metastasis; \\
\addlinespace[2pt]
\cline{2-3}
\addlinespace[2pt]
  & Desmoplasia & Pronounced desmoplastic reaction; growth of fibrous tissue; connective tissue increase; association with aggressive tumors; \\
\addlinespace[2pt]
\cline{2-3}
\addlinespace[2pt]
  & Tumor Budding & \makecell[l]{Small clusters of cancer cells at invasive front; individual cells at invasive front; sign of \\ aggressive tumor phenotype; correlated with metastasis;} \\
\addlinespace[2pt]
\cline{2-3}
\addlinespace[2pt]
  & Poorly Differentiated Tumor Cells & High-grade dedifferentiation; higher likelihood of metastasis; \\
\addlinespace[2pt]
\cline{2-3}
\addlinespace[2pt]
  & Hyperchromasia & Nuclei appear darker; excess DNA content; \\
\addlinespace[2pt]
\cline{2-3}
\addlinespace[2pt]
  & Increased TP53 Mutations & \makecell[l]{TP53 enrichment in high-CIN tumors; link to mitotic stress; TP53 malfunctions; increased mitotic \\ figures in histology; atypical nuclear features; increased nuclear size; irregular nuclear contours; \\ hyperchromasia; prominent nucleoli; genomic instability; altered cell cycle regulation; variety of cell types; \\ abnormal tumor structures;} \\
\addlinespace[2pt]
\cline{2-3}
\addlinespace[2pt]
  & Correlated Lauren's Intestinal Type & \makecell[l]{Well-formed glandular structures; intestinal epithelium resemblance; chronic gastritis initiation; \\ progression to atrophy; intestinal metaplasia; dysplasia; carcinoma development; common in high-incidence regions; \\ environmental factor association; diet-related; Helicobacter pylori infection;} \\
\addlinespace[2pt]
\cline{2-3}
\addlinespace[2pt]
  & Prominent Nucleoli & Prominent nucleoli; increased number of nucleoli; sign of heightened protein synthesis; rapid cell division indicator; \\
\addlinespace[2pt]
\cline{2-3}
\addlinespace[2pt]
  & Tumor Heterogeneity & \makecell[l]{CIN-induced genetic heterogeneity; RAS-driven proliferation of diverse cells; increased tumor complexity; \\ potential influence on drug resistance; enhancement of \\ metastatic potential;} \\

\bottomrule
\end{tabular}
}
\caption{Pathology concepts for MSI vs. Others.}
\label{tab:msi_vs_others_concepts}

\end{table*}

\begin{table*}[ht]
\centering
\resizebox{\linewidth}{!}{
\begin{tabular}{l l l}
\toprule
\textbf{Type} & \textbf{Concept} & \textbf{Description} \\ 
\midrule

\multirow{15}{*}{EBV} 
  & Lymphoepithelioma-like Histology & \makecell[l]{EBV-positive; lymphoepithelioma-like carcinoma; large sheets; syncytial clusters; undifferentiated cells; \\ prominent lymphoid infiltration; no glandular formation; non-keratinizing; vesicular nuclei; \\ prominent nucleoli; desmoplastic reaction;} \\
\addlinespace[2pt]
\cline{2-3}
\addlinespace[2pt]
  & Tumor Infiltrating Lymphocytes & Tumor-infiltrating lymphocytes; dispersed or clustered; infiltrating between cells or stromal; indicative of immune response; \\
\addlinespace[2pt]
\cline{2-3}
\addlinespace[2pt]
  & Intraepithelial Lymphocytosis & \makecell[l]{Intraepithelial lymphocytes; small, round; dense nuclei; disrupts architecture; associated with neoplastic epithelium; \\ stromal lymphoplasmacytic infiltration;} \\
\addlinespace[2pt]
\cline{2-3}
\addlinespace[2pt]
  & Stromal Lymphoplasmacytic Infiltration & \makecell[l]{Lymphocytes in stroma; plasma cells present; small cells with large nuclei; abundant basophilic cytoplasm; \\ interspersed infiltration; reactive changes; possible fibrosis or edema;} \\
\addlinespace[2pt]
\cline{2-3}
\addlinespace[2pt]
  & Syncytial trabecular pattern & Syncytial trabecular pattern; nested growth; cord-like structures; indistinct cell borders; interconnected net-like structure; \\
\addlinespace[2pt]
\cline{2-3}
\addlinespace[2pt]
  & Lace-like Pattern & Lace-like pattern; irregularly anastomosing tubules and cords; complex interconnected network; irregular net-like structure; \\
\addlinespace[2pt]
\cline{2-3}
\addlinespace[2pt]
  & Lymphoid Stroma & \makecell[l]{Lymphoid stroma infiltration; “lace-like” pattern; irregular tubules and cords; immune component in \\ microenvironment; variable lymphoid infiltration;} \\
\addlinespace[2pt]
\cline{2-3}
\addlinespace[2pt]
  & Invasion into the Submucosa & \makecell[l]{Invasion into submucosa; scattered cells to clusters; neoplastic cells breach muscularis \\ mucosae; lymphocytic response around cancer cells;} \\
\addlinespace[2pt]
\cline{2-3}
\addlinespace[2pt]
  & Poor Differentiation & Poorly differentiated adenocarcinomas; lacks specialized features; aggressive tumor; unformed glandular structures; infiltrating lymphoid stroma; \\
\addlinespace[2pt]
\cline{2-3}
\addlinespace[2pt]
  & Ulcered or saucer-like tumor & \makecell[l]{Central necrosis; ulceration with epithelial loss; robust inflammatory infiltrate; reactive \\ cellular changes; marginal roll at ulcer edges; increased vascularity; surrounding fibrosis;} \\

  \midrule

\multirow{12}{*}{Others} 
  & Increased Mitotic Activity & Increased mitotic rate; atypical mitotic figures; abnormal mitoses; high cellular proliferation; \\
\addlinespace[2pt]
\cline{2-3}
\addlinespace[2pt]
  & Nuclear Pleomorphism & Variation in nuclear size and shape; nuclei size disparity; irregular nuclear shapes; oval to highly irregular forms; \\
\addlinespace[2pt]
\cline{2-3}
\addlinespace[2pt]
  & Hyperchromasia & Nuclei appear darker; excess DNA content; \\
\addlinespace[2pt]
\cline{2-3}
\addlinespace[2pt]
  & Irregular Nuclear Contours & Uneven nuclear borders; indented nuclear contours; \\
\addlinespace[2pt]
\cline{2-3}
\addlinespace[2pt]
  & Prominent Nucleoli & Prominent nucleoli; increased number of nucleoli; sign of heightened protein synthesis; rapid cell division indicator; \\
\addlinespace[2pt]
\cline{2-3}
\addlinespace[2pt]
  & Chromatin Clumping & Irregular chromatin clumping; patchy nuclear appearance; \\
\addlinespace[2pt]
\cline{2-3}
\addlinespace[2pt]
  & Multipolar spindles & \makecell[l]{Multipolar spindles; asymmetric nuclear division; uneven genetic material distribution; \\ cells with abnormal nuclear shapes and sizes;} \\
\addlinespace[2pt]
\cline{2-3}
\addlinespace[2pt]
  & Tumor Budding & Tumor budding presence; aggressive tumor phenotype; correlated with metastasis; \\
\addlinespace[2pt]
\cline{2-3}
\addlinespace[2pt]
  & Lymphovascular Invasion & Tumor cells in lymphatic vessels; tumor cells in blood vessels; direct indication of metastasis; \\
\addlinespace[2pt]
\cline{2-3}
\addlinespace[2pt]
  & Desmoplasia & Pronounced desmoplastic reaction; growth of fibrous tissue; connective tissue increase; association with aggressive tumors; \\

\bottomrule
\end{tabular}
}
\caption{Pathology concepts for EBV vs. Others}
\label{tab:ebv_vs_others_concepts}

\end{table*}

\begin{table*}[ht]
\centering
\resizebox{\linewidth}{!}{
\begin{tabular}{l l l}
\toprule
\textbf{Type} & \textbf{Concept} & \textbf{Description} \\
\midrule
\multirow{10}{*}{Positive} 
  & HER2 Overexpression & Strong; complete membrane staining; indicative of HER2 positivity; \\
\addlinespace[2pt]
\cline{2-3}
\addlinespace[2pt]
  & High Tumor Cellularity & Densely packed cells; high nuclear-to-cytoplasmic ratio; scant stroma; ‘blue’ appearance from dense nuclear staining; \\
\addlinespace[2pt]
\cline{2-3}
\addlinespace[2pt]
  & Mitotic Figures & Numerous in aggressive tumors; cells in division; high proliferation rate; \\
\addlinespace[2pt]
\cline{2-3}
\addlinespace[2pt]
  & Necrosis & Dead cell areas; cell debris; lost tissue architecture; outpaced blood supply; \\
\addlinespace[2pt]
\cline{2-3}
\addlinespace[2pt]
  & Pleomorphism & Variation in size and shape of cells and nuclei; \\
\addlinespace[2pt]
\cline{2-3}
\addlinespace[2pt]
  & High Tumor-infiltrating Lymphocytes Levels & Inferred from H\&E sections; small, round, darkly stained nuclei; scant cytoplasm; \\
\addlinespace[2pt]
\cline{2-3}
\addlinespace[2pt]
  & Dense Clustering & Large, densely packed cellular areas on H\&E; high nuclear to cytoplasmic ratio; minimal stroma; \\
\addlinespace[2pt]
\cline{2-3}
\addlinespace[2pt]
  & Loss of E-Cadherin & Negative staining pattern; distinguishes lobular from ductal carcinoma; \\
\addlinespace[2pt]
\cline{2-3}
\addlinespace[2pt]
  & GCDFP-15 Positive & Cytoplasmic granular staining; secreted protein indicating apocrine differentiation; \\
\addlinespace[2pt]
\cline{2-3}
\addlinespace[2pt]
  & Nuclear Markers & \makecell[l]{High density of nuclei; ER, PR, Ki-67, p53 staining; \\ Ki-67 shows high proliferation index;} \\
\midrule
\multirow{10}{*}{Negative} 
  & HER2 Protein Regular & No membrane staining or $\leq$10\% staining; partial membrane staining in $\geq$10\% of cells; \\
\addlinespace[2pt]
\cline{2-3}
\addlinespace[2pt]
  & Hormone Receptor Negative & No nuclear staining for ER/PR; consistent absence across cancer cells; uniform lack of staining; \\
\addlinespace[2pt]
\cline{2-3}
\addlinespace[2pt]
  & ER Negative & No nuclear staining; antibodies against ER don't bind; \\
\addlinespace[2pt]
\cline{2-3}
\addlinespace[2pt]
  & PR Negative & No nuclear staining; antibodies against PR don't bind; \\
\addlinespace[2pt]
\cline{2-3}
\addlinespace[2pt]
  & K67 Proteins & Nuclear staining; marks cell proliferation; absent in non-proliferative cells; \\
\addlinespace[2pt]
\cline{2-3}
\addlinespace[2pt]
  & DDR (DNA damage response) Effective & \makecell[l]{Lack/reduced expression; indicative of defective DNA repair; \\ susceptibility to DDR inhibitors;} \\
\addlinespace[2pt]
\cline{2-3}
\addlinespace[2pt]
  & Blood Vessel Density & CD31 or CD34 positive staining; lines blood vessels; increased density indicates active angiogenesis; \\
\addlinespace[2pt]
\cline{2-3}
\addlinespace[2pt]
  & Increased EMT (epithelial-mesenchymal transition) & Increased expression; suggestive of metastasis facilitation; \\
\addlinespace[2pt]
\cline{2-3}
\addlinespace[2pt]
  & Tumor Cell Invasion & Increased expression; indicates invasive potential; \\
\addlinespace[2pt]
\cline{2-3}
\addlinespace[2pt]
  & Vimentin Positive & Cytoplasmic staining; mesenchymal cell cytoskeletal component; \\
\midrule
\multirow{10}{*}{Equivocal} 
  & IHC Score 2+ & No staining; faint staining; $\leq$ 10\% tumor cells; \\
\addlinespace[2pt]
\cline{2-3}
\addlinespace[2pt]
  & HER2 Low Expression & Faint staining; barely perceptible; $\geq$10\% tumor cells; \\
\addlinespace[2pt]
\cline{2-3}
\addlinespace[2pt]
  & HER2 Ultra-Low Expression & \makecell[l]{Weak to moderate staining; complete membrane; $\geq$ 10\% tumor cells; \\ Strong staining; complete membrane; $\leq$10\% tumor cells;} \\
\addlinespace[2pt]
\cline{2-3}
\addlinespace[2pt]
  & Heterogeneity & Variable HER2 expression; within the same tumor; challenging determination; \\
\addlinespace[2pt]
\cline{2-3}
\addlinespace[2pt]
  & Variable Staining Intensity & Variable intensity; across tumor areas; some regions stronger than others; \\
\addlinespace[2pt]
\cline{2-3}
\addlinespace[2pt]
  & Identified Invasive Tumor & Spread into tissues; potentially worse prognosis; beyond ducts/lobules; \\
\addlinespace[2pt]
\cline{2-3}
\addlinespace[2pt]
  & Moderate Tumor Proliferation & Lower than HER2-positive; higher than HER2-negative; complete but moderate membrane staining; \\
\addlinespace[2pt]
\cline{2-3}
\addlinespace[2pt]
  & Moderate Tumor Grading & Moderate uniformity; variable intensity and completeness; across tumor population; \\
\addlinespace[2pt]
\cline{2-3}
\addlinespace[2pt]
  & Metastatic Focus & Clusters of atypical cells; different from lymphoid cells; IHC markers highlight cancer cells; \\
\addlinespace[2pt]
\cline{2-3}
\addlinespace[2pt]
  & Moderate Residual Cancer Burden (RCB) & $\geq$10\% tumor cells; weak/moderate intensity; \\
\bottomrule
\end{tabular}
}
\caption{Pathology concepts for Positive vs. Negative vs. Equivocal in BRCA HER2 prediction task}
\label{tab:her2_concepts}

\end{table*}

\end{document}